\documentclass[acmlarge]{acmart}

\usepackage{booktabs} 
\usepackage{dirtytalk} 
\usepackage[inline]{enumitem} 
\usepackage{hyperref} 
\usepackage[]{todonotes} 
\usepackage{longtable} 
\usepackage{multirow}
\usepackage{svg}
\usepackage[utf8]{inputenc}

\newcommand{\eg}{\textit{e.g.},~}
\newcommand{\ie}{\textit{i.e.},~}

\setcopyright{acmcopyright}
\copyrightyear{2022}
\acmYear{2022}

\acmDOI{XXXXXXX.XXXXXXX}

\acmJournal{CSUR}
\acmVolume{00}
\acmNumber{0}
\acmArticle{000}
\acmMonth{0}

\begin{document}
\title[Debiasing Methods for Fairer Neural Models in Vision and Language Research: A Survey]{Debiasing Methods for Fairer Neural Models\\ in Vision and Language Research: A Survey}

\author{Otavio~Parraga}
\authornote{The authors contributed equally to this research.}
\author{Martin~D.~More}
\authornotemark[1]
\author{Christian~M.~Oliveira}
\authornotemark[1]
\author{Nathan~S.~Gavenski}
\authornotemark[1]
\author{Lucas~S.~Kupssinskü}
\author{Adilson~Medronha}
\author{Luis~V.~Moura}
\author{Gabriel~S.~Simões}
\author{Rodrigo~C.~Barros}
\affiliation{%
  \institution{Machine Learning Theory and Applications (MALTA) Lab, PUCRS}
  \country{Brazil}}

\renewcommand{\shortauthors}{Parraga~et~al.}

\begin{abstract}
Despite being responsible for state-of-the-art results in several computer vision and natural language processing tasks, neural networks have faced harsh criticism due to some of their current shortcomings.
One of them is that neural networks are correlation machines prone to model biases within the data instead of focusing on actual useful causal relationships.
This problem is particularly serious in application domains affected by aspects such as race, gender, and age.
To prevent models from incurring on unfair decision-making, the AI community has concentrated efforts in correcting algorithmic biases, giving rise to the research area now widely known as \textit{fairness in AI}.
In this survey paper, we provide an in-depth overview of the main debiasing methods for fairness-aware neural networks in the context of vision and language research.
We propose a novel taxonomy to better organize the literature on debiasing methods for fairness, and we discuss the current challenges, trends, and important future work directions for the interested researcher and practitioner. 
\end{abstract}



\maketitle

\section{Introduction}

Deep Learning is a subfield of Machine Learning (ML) that leverages the capabilities of artificial neural networks to automatically learn from data.
These networks are fully-differentiable computational graphs optimized via gradient descent to learn representations from raw data~\cite{bengio2013representation}, currently being the most efficient and effective data-oriented strategy to perform several Computer Vision (CV) and Natural Language Processing (NLP) tasks.

Despite producing exciting results, neural models have faced harsh criticism due to some of their current shortcomings.
One of the main criticisms is that, since neural networks are correlation-based approaches, models often learn the influence of confounding factors that are present in the data instead of causal relationships~\cite{adeli2019bias}.
This problem is exacerbated in application domains affected by sensitive (or protected) features, such as demographic information.
For instance, race, gender, and age may confound the training process because the distribution of labels is often skewed, even if the intrinsic properties of interest are not related to them.
Thus, the presence of confounding factors may result in models that are biased towards specific distributions, potentially exhibiting a severe drop in performance in unseen data or prioritizing certain subgroups within a distribution.

Given the widespread usage of neural models, one must consider the impact of their respective automated decisions.
Applications such as product recommendations and automatic game-playing are considered to be low-stakes since biased behaviors do not significantly impact underrepresented groups in society.
However, automated decisions in dating, hiring~\cite{Bogen2018,Cohen2019}, and loan management~\cite{Srinivasan2019} software present considerably higher stakes, given that they may influence and perpetuate unfair economic and social disparities between groups.

To address and regulate algorithmic usage in high-stakes decision making, several governmental entities have proposed the creation of stronger laws requiring more transparency in automated decision-making.
For example, the European Union GDPR~\cite{gpdr-art22} states that people should have the right to obtain an explanation of the decision reached by automated systems.
Indeed, explainability seems to play a crucial role towards achieving trustworthy AI, \ie systems that are lawful, technically robust, bias-resilient, and ethically adherent.
Since neural networks are black-box models that require external tools to extract explanations~\cite{guidotti2018survey}, researchers and practitioners often ignore such tools, allowing the design and optimization of unfair models.

To prevent learning models from perpetuating the biases present in the data and producing unfair decisions in automated decision-making, the Artificial Intelligence (AI) community has concentrated efforts in correcting algorithmic biases, giving rise to the research area now widely known as \textbf{fairness} in AI.
When considering the process of automated decision-making, the term \textit{fairness} refers to the ``absence of any prejudice or favoritism toward an individual or a group based on their inherent or acquired characteristics''~\cite{mehrab2019survey}.
Fairness in AI is a relatively new research area.
Initially, fairness papers were mostly submitted to workshops focused on data privacy, with occasional papers appearing in main proceedings.
Starting in 2014, we see the appearance of several workshops and conferences specialized in fairness, accountability, and transparency in machine learning.
Notable examples include FAT/ML (2014-2018), AIES (2018-present), FAccT (formerly FAT*, 2018-present), and FATES (2019-present).
As the research field matured, we see a noticeable increase in the number of fairness papers that appeared in the main proceedings of renowned AI and Machine Learning conferences, such as AAAI, NeurIPS, ICML, ICLR, and many others.
These conferences now also occasionally host tutorials and workshops dedicated mainly for fairness-related approaches in data-driven learning, and some conferences, such as NeurIPS, are experimenting with implementing fairness awareness protocols, such as
including a ``Broader Impact'' section, covering the ethical aspects of the algorithms being proposed, and 
proposing a list of best practices for responsible machine learning research~\cite{neurips2021paperchecklist1, neurips2021paperchecklist2}.
This historical summary shows that the research community in general is starting to focus not only on improving the performance of algorithms, but also in creating fairer ones.

Given the relevancy and recent proliferation of fairness concerns in automated decision-making, our goal in this survey is to provide an overview of the progress and current state of the art in neural approaches for fairness and bias mitigation in AI.
We focus on vision and language research since these data modalities and their intersection encompass the majority of neural networks research.

Several surveys of fairness in AI already exist, each covering different aspects of the research area.
For instance, the work of \citet{mehrab2019survey} focuses on fairness in ML while extensively detailing fairness definitions and types of biases, but offers a quick analysis of debiasing methods in ML in general.
A similar content and structure can be found in~\cite{caton2020fairness,dunkelau2019fairness}, where a broad analysis is performed for different ML tasks and algorithms.
The work of \citet{le2022survey} goes in a different direction by focusing exclusively on analyzing tabular datasets and their usages for bias mitigation.
\citet{Tian2022}, in turn, cover fairness exclusively for image data.
Finally, there are also several survey papers that extensively cover bias and fairness solely within the scope of NLP~\cite{meade2021empirical, delobelle2021measuring, garrido2021survey, blodgett2020language, bansal2022survey}, be it only regarding pre-trained language models (LMs)~\cite{meade2021empirical, delobelle2021measuring}, or exclusively for deep learning~\cite{garrido2021survey}.

In contrast with the aforementioned studies, this work focuses on an in-depth analysis of neural-based methods for debiasing in the context of the main unstructured data types, namely visual and textual data and their intersection (\ie multimodal tasks).
By focusing on debiasing approaches, we offer a new taxonomy for properly categorizing those methods while following the detailed definitions of fairness and bias proposed by~\citet{mehrab2019survey}, which are now well-accepted and widely used by the research community. 

This work surveys $95$ debiasing methods exclusively in the context of neural networks for vision and language. The only work to survey a similar amount of methods is the one by \citet{caton2020fairness}, which reviews a total of $86$ debiasing methods, though across all ML research. Still, there is only one paper in the intersection of this work and \cite{caton2020fairness}, which is the work from \citet{edwards2016censoring}. The remaining survey papers on fairness review a much smaller set of methods for debiasing. \citet{dunkelau2019fairness} survey $24$ debiasing methods, and once again its intersection with our work is only a single paper, \citep{edwards2016censoring}. Finally, the outstanding survey by \citet{mehrab2019survey} only presents $13$ debiasing methods, none of them falling under our criterion for acceptance, namely being a method for making neural models fairer in the context of vision and language research.
 
\section{Scope and Organization}

In this section, we detail the scope and organization of this paper. First, we contextualize fairness and its relationship with bias within neural network research in Section~\ref{sec:bias-and-fairness}.
\textit{Bias} is an overused word in ML research and has several different meanings according to the context where it is used.
Learning requires methods to incorporate \textit{inductive biases}, which are preferences towards choosing/modeling certain solutions instead of others.
In the context of neural networks, \textit{biases} are also attribute-free parameters that shift the model according to prior information.
The correlation-based approach implemented in most ML methods, neural networks being no exception, may capture relationships that lead to \textit{biased} solutions, \ie a model whose behavior may be undesired because of spurious correlations that were captured during training.
In this paper, when we talk about \textit{biases} we are not talking about the attribute-free parameters in neural networks. 
We are also rarely talking about inductive bias, since algorithmic biases are seldom the cause of \textit{fairness} problems, though that may also happen.

Hence, most of the time we mention the word \textit{bias} we mean unintended behavior resulting from correlation-based processing that ignores further context not explicit in the data.
In practice, we assume that exploiting correlations is the own nature of ML algorithms, and it is the combination of biased samples and correlation identification that generates the problem of models that are biased towards undesired behavior.

Fairness is also a term with many proposed definitions.
We follow the specialized literature and situate (un)fairness as a direct consequence of capturing spurious correlations during training, as long as those correlations result in the ``\textit{favoritism toward an individual or a group based on their inherent or acquired characteristics}''~\cite{mehrab2019survey}.
By situating problems with fairness as a direct consequence of existing correlation-based biases, we show that most bias-mitigating strategies can be used to improve the fairness of automatic systems, and therefore we establish the link between the areas of bias mitigation and fairness awareness in Section~\ref{sec:bias-and-fairness}.

Next, we comment on the main metrics and evaluation measures for assessing the level of fairness of a neural model in the context of those applications in Section~\ref{sec:metrics}.
We review both application-specific and general-purpose measures, pointing to their proper use, applicability, and known limitations.
The core of this survey, however, is the critical analysis and discussion of several debiasing methods for neural models in the context of image and language processing, which we present in Section~\ref{sec:debias-methods}.
Computer vision and natural language processing are arguably the two most important research areas in AI nowadays, given the amount of content produced that falls into these categories and their intersection.
For instance, in 2021 approximately $1.4$ trillion digital photographs were taken worldwide\footnote{https://blog.mylio.com/how-many-photos-taken-in-2022/} while Twitter users posted approximately $200$ billion tweets\footnote{https://www.internetlivestats.com/twitter-statistics/}.

Finally, we list the current fairness challenges in neural models, highlighting trends and important future research directions in Section~\ref{sec:challenges-future-directions}. 
We also specifically envision the challenges of fairness awareness in the so-called Foundation Models~\cite{bommasani2021opportunities}, which are extremely-large neural networks trained over massive amounts of data.
The challenge of addressing fairness unawareness in models that were trained self-supervisedly is probably one of the main factors that prevent such models from being fully open-sourced, given that this category of neural models has the potential of being applied and adapted to several tasks and biases may be perpetuated or even potentialized if not considered and properly addressed, resulting in unfair decision-making.

\section{Bias and Fairness}
\label{sec:bias-and-fairness}

The word \textit{bias} is used in ML in many distinct contexts.
\citet{mitchell1980need} defines bias as \say{any basis for choosing one generalization over another, other than strict consistency with the instances}.
This definition encompasses biases that are inherent to the learning task and are unavoidable, the so-called inductive biases.
In convolutional neural networks, for instance, the implemented inductive bias explores the fact that the input data often has spatial coherence and that higher level features are translation and rotation invariant.
Another common use of the word \textit{bias} is to refer to the parameters of the neural network that are free w.r.t. the input features.
In this survey, we will focus on biases that are not mandatory in the learning process and that can often skew the results of the model in undesired ways, with the possibility of causing social harm. 

Since artificial neural networks are essentially data-driven methods, data is the main source of unwanted and potentially avoidable bias.
In an ideal scenario, data should be complete, correct, and representative of a population of interest.
However, it is often the case that the sampling scheme, the measurement, or the representation of the data is biased towards a group of subjects. 
\citet{olteanu2019social} call this scenario \textit{data bias}.

It is also common to use datasets that follow a given distribution and expect them to generalize to another (perhaps slightly distinct) one.
A facial attribute recognition system could be trained using the CelebA dataset~\citep{liu2015deep}, but when the model is deployed to recognize facial features in the wild we can expect a performance drop given that the target population is not a group of celebrities.
This type of bias arises with the distinction of distributions from training and production settings.
Even assuming that the dataset is representative of the population that we are interested in, we are still prone to create biased models. 
Take the scenario of NLP systems that are trained in very large corpora of text.
The word embeddings created by these systems incorporate stereotypes that are found in the text, such as associating the word \textit{engineering} more frequently with men than women~\cite{suresh2019framework}.

Many studies searched for ways to categorize bias following different taxonomies, each based on a different system and set of assumptions.
The work of \citet{mehrab2019survey} divides biases based on the relationship among three key entities that constantly interact with each other: user, algorithm, and data.
Each entity will aggregate a group of possible biases that may occur in interactions within the ML life cycle, which may affect another entity in the process.
The relationship between data and learning algorithm clearly allows for the identification of outcome disparities and biases. 
Historical, social, cultural, and economic factors, as well as cognitive human biases, may affect all the process of collecting data, biasing samples and entire datasets without being easily detected.
The relationship between algorithm and user can also be a source for potential biases to arise, specially when the algorithm is responsible for guiding the human behaviour to specific patterns~\cite{Mengnan2021}.

In the study of \citet{suresh2019framework}, the types of biases are defined according to the ML life cycle.
It is established that unintended biases can be inserted within the data (be it a problem of data generation, representation, or measurement) or by model building and implementation (due to the learning process, evaluation procedure, aggregation of models, or the deployment in an environment where the concepts that were modelled do not apply).
Although this definition is more straightforward than the one in \cite{mehrab2019survey}, it lacks the capability of differentiating that data biases such as \textit{content production} and \textit{aggregation} are created by user-data interaction and data-algorithm interaction, respectively.

We can also view the bias phenomena as an \textit{origin and consequence} framework~\cite{shah2020predictive}.
More focused on NLP applications, \citet{shah2020predictive} define four origins for biases in source data: over-amplification, semantic, selection, and label; and two consequences in the outcomes: outcome disparity and error disparity.
By using a standard supervised pipeline of NLP, the authors attribute each bias origin to a different step, from the embedding phase where the semantics are condensed into a dense vectorized representation to the possibility of over-amplification due to the combination of the learning algorithm and the data itself or the under-representative selection of instances.
The consequences are defined by two disparities obtained when analyzing the output distributions.

\begin{figure}[!tpb]
    \centering
    \includegraphics[width=0.65\textwidth]{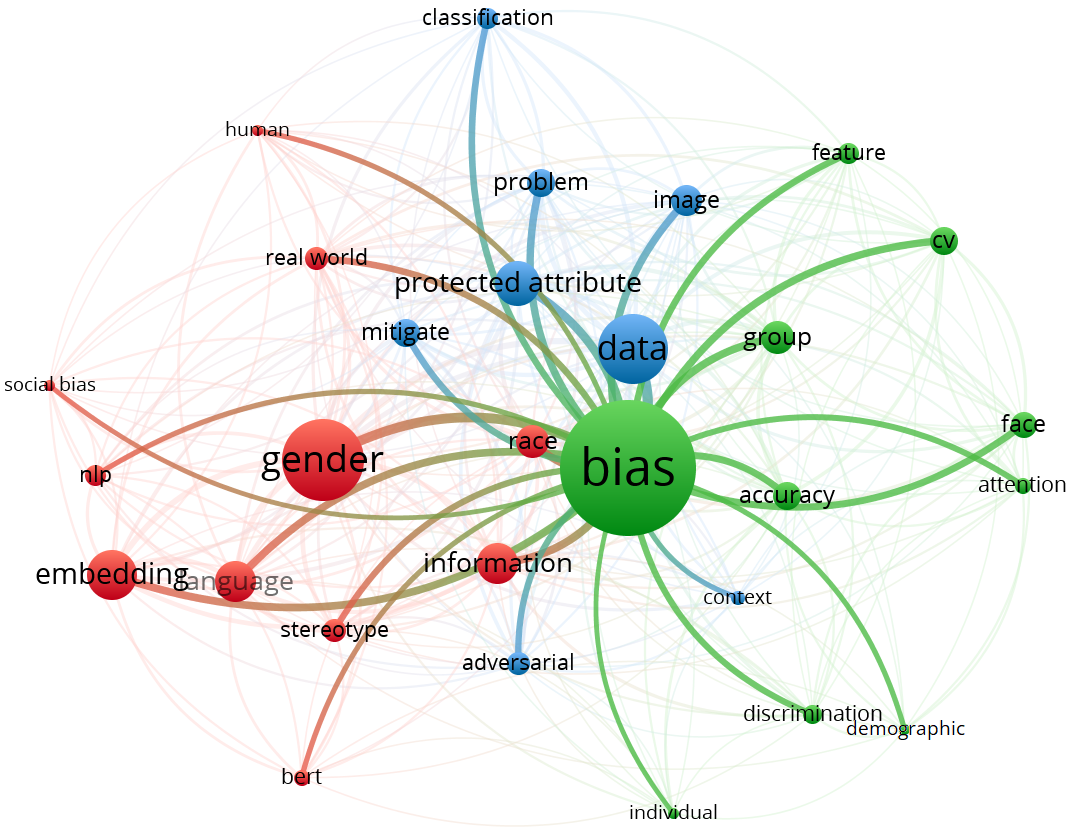}
    \caption{Network of terms found in the abstracts of the surveyed papers. The term \textit{bias} and its relationship with other terms are highlighted. Created with VOSViewer~\citep{perianer2016bibliometric}.}
   \label{fig:wordcloud}
\end{figure}

Mitigating bias in neural models is not a trivial task.
Despite the discussion regarding adopted taxonomies and a strict definition of bias~\cite{shen2022unintended,olteanu2019social}, there is a consensus that biased models can cause societal harm and that researchers and practitioners must be vigilant and employ tools to detect and mitigate this problem. 
A point should be made that, although humans have an unmatched capacity to learn from data and to generalize to different contexts, we too are prone to biases and to being unfair.
Nonetheless, the problem is exacerbated by neural models since they can work on a larger scale and reinforce negative feedback loops in society~\cite{Kusner2017}.

When biases in the learned models are detrimental to specific groups, often defined by sensitive attributes, we have models that are \textit{unfair}. 
The question of fairness is specially relevant when automated decisions exacerbate prejudicial behavior against socially-vulnerable groups or when it promotes favoritism towards a specific demography, gender, ethnicity, or race \cite{Mengnan2021}. 
Model fairness leverages moral questions about how we collect, validate, and use data, especially now with models getting larger and datasets reaching massive sizes~\cite{brown2020language}.
We cannot disregard group representation nor oversee undesired outcomes in trained models anymore.

When researchers and practitioners focus solely on traditional performance metrics such as accuracy, a trained model can propagate stereotypes and biases present in the training data.
While general fairness is a well-defined concept, the definition of what is the best measure to capture fairness is still subject of debate~\cite{caton2020fairness}.
One should be careful since distinct definitions of fairness derived from the use of different metrics have mutual interactions~\cite{kleinberg2018algorithmic}, \eg when optimizing for group fairness it is possible to make the situation worse for individual fairness.
These scenarios are presented with greater detail in Section~\ref{sec:metrics}.

We illustrate in Figure~\ref{fig:wordcloud} the landscape of fairness research in neural networks for vision and language with a network of terms found in the abstracts of all surveyed papers.
In such visualization, the node area is proportional to the number of occurrences of the term within the abstracts.
Note that \textit{bias} and \textit{data} are the most used terms, considering that the main source of bias in ML applications is the data and their interaction with users and algorithms.
The modalities that we are focusing on, nodes \textit{cv} and \textit{nlp}, as well as their intersection are also commonly present in the metadata or through proxy terms such as \textit{language}, \textit{image} and \textit{embedding}. 
We can also see that the term \textit{bias} appears in association with attributes such as \textit{gender}, and \textit{race}, which are indeed often associated with \textit{subgroups} or \textit{individuals} that are harmed by unfair decisions.
Other terms that relate to sensitive attributes are \textit{demographic}, \textit{ethnicity}, and \textit{social bias}.

Fairness is not achieved solely by guaranteeing the same outcome for distinct groups.
\citet{mehrab2019survey} present the case where a seemingly unfair difference in annual income between men and women in the UCI Adult dataset can be explained in terms of working hours.
In that dataset, men on average work more hours per week than women.
Working hours is thus a legitimate attribute that explains the difference in annual income and should not be considered an issue in the context of fairness.

To promote fairness, a special subset of available features are regarded as \textit{sensitive} or \textit{protected}.
They help define subgroups that are considered underprivileged in society.
An example of protected attribute is gender, and there are several examples of ML systems presenting error rates skewed towards a particular gender category~\cite{Buolamwini2018gender,guo2021detecting,rose2021bias}.
The definition of the the subset of attributes that are sensitive is not an entirely technical issue.
Although there is certain agreement that ethnicity should be protected in some applications, it could be the case that other attributes such as zip code serve as a proxy for ethnicity and should also be protected.
Furthermore, despite being a protected attribute, ethnicity matters in certain applications, \eg medical diagnosis where genetic predisposition is a determining factor.
In practice, any attribute that society perceives as a potential defining factor for aggregating underprivileged people is a potential sensitive attribute.

Fairness can also be stratified regarding groups, subgroups, or individuals.
Fairness in groups can be defined through the \textit{Demographic Parity} concept, which states that the likelihood of an outcome should be the same for a person in or out of the protected group. 
On the other hand, there is also the \textit{Counterfactual Fairness} definition, which states that inferences over a single individual should be kept unaltered in a hypothetical counterfactual world where the same individual belongs to a distinct group. 
Note that satisfying every definition of fairness at the same time is not feasible in real-world applications~\cite{lambrecht2019algorithmic}, and we exemplify that in Section~\ref{sec:metrics}.

Issues of model fairness are often intertwined with biases in the ML pipeline. 
Datasets such as IJB-A~\cite{klare2012facerecognition} and Adience~\cite{levi2015agegender} do not have proper representation of gender and skin color~\cite{Buolamwini2018gender}. 
Models trained in biased datasets will often produce worse results for individuals in underrepresented groups.
They may, for instance, misclassify Black women with a higher probability than lighter-skinned men.
In that situation, the unfairness of the model can be traced back to its origin in a biased dataset.

When the underprivileged group is defined by a combination of sensitive attributes, and the dynamics of individuals in this group is considerably distinct than when considering one sensitive attribute at a time, we have \textit{intersectional} biases~\cite{Buolamwini2018gender}.
Unfairness caused by intersectional biases are harder to detect and to prevent because of intricate relationships among the attributes~\cite{guo2021detecting,rose2021bias}. 

Some applications of ML models are considered to be of low stakes, having a limited impact in society well-fare and errors are considered cheap. 
On the other hand, ML models are also being used in significant higher-stake applications, where a single error in inference is expensive and can negatively impact individuals and groups. 
Examples of these applications are facial recognition systems~\cite{Buolamwini2018gender}, criminal assessment~\cite{angwin2016compas}, and occupation classification~\cite{biasinbios}. 
Fairness is of the upmost importance in high-stake applications.
ML models should be subject to scrutiny as they are part of an entire ecosystem that influences different social groups, and can perpetuate harmful concepts and stereotypes when left unchecked.

When optimizing for model fairness, it is necessary to keep in mind that the level of data aggregation can also be a source of confusion. 
This type of problem is defined as Simpson's Paradox~\cite{blyth1972simpson}, which is the phenomenon in which the same dataset can lead to opposite trends by changing how the data is aggregated, always happening when the aggregated data hide conditional variables. 
Regarding neural networks, most features are learned by searching for correlations in data when trying to optimize an objective function, so the undesired lack of fairness may be a product of seeking good performance, creating a kind of performance-accuracy trade-off. 

Most research in fairness cover the ML field without focusing on deep learning~\cite{mehrab2019survey,suresh2019framework,caton2020fairness}. 
As a consequence, lack of fairness in multimodal tasks (\eg text-to-image systems) are less understood than in more traditional classification tasks. 
Another specificity in neural-network research is the recent rise of Foundational Models (FMs)~\cite{bommasani2021opportunities}. 
Biases in those models are challenging to identify, and the own nature of FMs as unfinished models that need to be further adapted makes them a perfect fit for the problem of bias propagation and exacerbation within many downstream tasks. 
We further discuss this challenge in Section~\ref{sec:challenges-future-directions}.

As neural models are being increasingly deployed to real-world applications, the interest on fairness grows both in academia and industry.
While there are countries that are starting to regulate some aspects of fairness in ML models and automated decision-making~\cite{bird2019fairnessaware}, there is much to be done to achieve a unified framework that the research community can agree on and that practitioners can apply when developing fairer applications.
Among the requirements for the fairness research to provide practical impact in society, there should be effective methods that allow biases to be mitigated and underprivileged groups to be protected without a significant compromise in terms of efficiency and effectiveness. We categorize and review those methods in Section~\ref{sec:debias-methods}, while also discussing their advantages and shortcomings.

\section{Metrics}
\label{sec:metrics}

This section describes the most relevant metrics to measure bias and fairness of neural models in CV and NLP, though all metrics presented in Section~\ref{classif} apply to any type of classification problem.

\subsection{Metrics for Classification}
\label{classif}
Classification is a traditional problem in machine learning and one of the most common tasks in computer vision.
During the training process, given a dataset $D=\{(x_i, y_i)\}_{i=1}^{N}$ the algorithm will learn how to correlate a set of inputs $X$ to a set of corresponding target variables $Y$.
The most common metric to evaluate classification models is \textit{accuracy}, the rate of correctly-classified objects.
Consider the confusion matrix of a binary classification problem, comprised of True Positives (Negatives) as the amount of objects correctly classified from the positive (negative) class, and False Positives (Negatives) as the amount of misclassified objects from the positive (negative) class.
With these same terms, we can derive three other widely-used metrics: \textit{precision}, \textit{recall}, and \textit{F1-Score}.

Even though those metrics are well-establish for evaluating the general performance of a learning model, they are not well-suited for fairness evaluation.
For example, consider a model trained on a dataset containing two groups, $A$ and $B$, where $A$ is privileged in society while $B$ is marginalized.
Societal biases have affected model performance, leading it to perform better on $A$ than on $B$.
Increasing accuracy using optimization techniques does not necessarily improve fairness.
It is possible the model gets even better at classifying $A$ and slightly worse on $B$, leading to an even unfairer model.
Similarly, precision, recall, and F1-score alone cannot guarantee model fairness, nor properly evaluate its nuances.
With this in mind, fairness-specific metrics have been proposed for classification models.
These metrics have been mainly divided into two categories: group fairness, which requires the average output of different demographic groups to be similar~\citep{Dwork2018}; and individual fairness, which requires that individuals who are similar in respect to a specific task have similar probability distributions on classification outcomes, regardless of the demographic group they are part of~\citep{Dwork2018, Gajane2017}.

\subsubsection{Group Fairness}
In group fairness, examples are grouped according to a particular protected attribute, and statistics about model predictions are calculated for each group and compared across them~\citep{Du2021}.
Next, we describe the most used group fairness metrics. 

\textbf{Demographic/Statistical Parity} states that the average algorithmic decision should be similar across different groups: $p(\hat{y}=c_k \mid z=p) \sim p(\hat{y}=c_k \mid z=u)$, where $\hat{y} = c_k$ is a predicted class and $z$ refers to a protected attribute such as race, gender, and age, in which $p$ and $u$ indicate privileged and underprivileged groups, respectively.
This metric does not depend on ground-truth labels, so it is especially useful when this information is unavailable.
When we are aware that historical biases may have skewed the data, demographic parity may be used in conjunction with a bias mitigation strategy to verify whether the model has learned those biases. 

\textbf{Equality of Opportunity} states that the true positive rate for individuals who qualify for the desirable/positive outcome ($y=1$) should be similar across different groups: $p(\hat{y}=1 \mid z=p, y=1) \sim p(\hat{y}=1 \mid z=u, y=1)$ where $y$ is the ground truth label.
This metric considers that different groups may have different distributions, which leads to one of its main criticisms: it does not consider the effect of discrimination given the protected attributes.

\textbf{Equality of Odds} states that both the true positive rate for individuals who qualify for the desirable/positive outcome and the false positive rate for individuals who do not qualify for the desirable/positive outcome should be similar across different groups.
The true positive rate is computed as the equality of opportunity, while the false positive rate is computed as $p(\hat{y}=1 \mid z=p, y=0) \sim p(\hat{y}=1 \mid z=u, y=0)$.
Equality of odds is a more restrictive variation of equality of opportunity, but the same criticisms of the former still hold for this metric. 

\textbf{Conditional Statistical Parity} states that different groups should have similar probability of being assigned to a positive outcome given that the individuals satisfy a set of legitimate factors $L$: $p(\hat{y}=1 \mid L=1, z=p) \sim p(\hat{y}=1 \mid L=1, z=u)$.
The set of factors $L$ is task-dependent and should be modeled accordingly. 

\textbf{Treatment Equality} states that the ratio of false negatives and false positives should be similar across different groups: $(FN/FP \mid y=1) \sim (FN/FP \mid y=0)$.
This metric may be used as a policy lever while optimizing for other metrics.
If a specific group has a higher ratio of false negatives in order to satisfy another metric, then it is actually treating these groups differently.

\textbf{Overall Accuracy Equality} states that accuracy should be similar for different groups:  $((TP+TN)/(TP+FP+TN+FN)\mid z=p) \sim ((TP+TN)/(TP+FP+TN+FN)\mid z=u)$.
This metric thus implies that true negatives are as desirable as true positives, being quite uncommonly-used for that reason.

\textbf{Predictive Parity} is similar to Overall Accuracy Equality, except that it considers that different groups should have similar precision instead of accuracy: $((TP)/(TP+FP)\mid z=p) \sim ((TP)/(TP+FP)\mid z=u)$.
Mathematically, a model with equal precision for different groups will also have equal false discovery rate: $((FP)/(TP+FP)\mid z=p) \sim ((FP)/(TP+FP)\mid z=u)$.

\textbf{Right for the Right Reasons}: Many classification errors occur due to the model looking at the wrong evidence.
It can happen based on any contextual evidence, \eg a multimodal model ignoring the input image when asked ``What color is the banana?'' since the most likely answer is ``yellow''. 
To compute this measurement quantitatively for image-based models, \citet{hendricks2018women} rely on two visual explanation techniques: Grad-CAM~\citep{ramprasaath2016gradcam} and saliency maps generated by occluding image regions in a sliding window fashion. 
In their work, they use the MS-COCO ground-truth person segmentation to evaluate whether the Grad-CAM from the model overlaps with the correct areas of the image when classifying a given class.
As this metric evaluates how the model gets its prediction, it is aligned with the current demand for more explainable AI.

\subsubsection{Individual Fairness}
Group fairness mainly ignores all information of the objects except for protected attribute $z$.
That strategy might hide unfairness, and we exemplify a scenario in which that happens as follows.
Suppose a model assigns a positive score to the same fraction of male and female applicants.
However, assume male applicants were chosen randomly whereas female applicants were chosen based on reference attribute values.
\textit{Demographic parity} will state that the model is actually fair, despite the discrepancy on how the applications were processed based on gender.
The notion of individual fairness arises to deal with such a discrepancy. 

\textbf{Fairness through Awareness} states that similar individuals should be treated similarly, where subject similarity is a task-specific metric.
Consider two \textit{similar} individuals, where the only major difference between them is that one is male and the other female.
Since they are similar, individual fairness states that they must follow the same distribution, generating the same output.
This strategy has already proven to be more efficient than \textit{Fairness through Unawareness}, which relied on the assumption that an algorithm would be fair if none of the sensitive attributes were explicitly used in the predictive process. 
Unfortunately, building a similarity score considering the different types of attributes is the biggest obstacle for putting this metric into practice. 

\textbf{Counterfactual fairness} states that an individual and its counterfactual copy whose only difference is the protected attribute value should have the same outcome.
While fairness through awareness finds similar individuals through task-specific measures, this metric generates synthetic copies of a counterpart instance (\eg female if the original instance is male in a gender-fair evaluation).
It also takes into account that several features might be dependant on the protected attribute and, therefore, should also be altered accordingly.

\subsubsection{Critical Analysis of the Classification Metrics}
Suppose we want to train a model to admit students to a given University with $30$ openings, and we have two groups of people: $A$, which comprises $70$ students, and $B$, with $30$.
In addition, assume $A$ is privileged in society whereas $B$ is part of a marginalized group.
Historical biases led students of group $B$ to have, on average, lower grades than group $A$, even though both groups contain students with variable performance in terms of grades.
Finally, assume $40$ students from group $A$ and $12$ from group $B$ reach the \textit{desirable} grade level to be admitted to this University. 
We discuss two different learning models and how the proposed metrics measure what is happening.
This discussion applies not only to neural networks but to any learning model and classification task in machine learning.

\textbf{Approach \#1}.
Assume a model was trained to achieve similar acceptance rates for both groups.
It selects $21$ students from group $A$ and $9$ from group $B$.
This model satisfies the \textit{Demographic Parity} criterion, since the percentage of accepted students was the same for both groups: $30\%$. However, if the selection were made randomly in either $A$ or $B$ instead of considering the grades, this metric would still evaluate this model as \textit{fair}.
On the other hand, this model does not satisfy the \textit{Equality of Opportunity} criterion: even if the selection process were made entirely based on each student's grades, the rate of acceptance considering only the students with desirable grades was higher for group $B$ ($75\%$) than for group A ($53\%$). 

\textbf{Approach \#2}.
Assume a second model was trained to achieve a similar acceptance rate in both groups, but this time only considering the students with desirable grades.
As a result, $23$ and $7$ students were accepted from groups $A$ and $B$, respectively.
In this scenario, \textit{Equality of Opportunity} is satisfied ($58\%$ for both groups), while \textit{Demographic Parity} is not ($33\%$ for group $A$ and $23\%$ for group $B$).
The problem here is that the historical biases that led to the disproportion regarding the grades for each group are not taken into account.
Therefore, this strategy alone will perpetuate historical biases that are currently present in society. 

It is easy to see that one cannot satisfy all fairness metrics without a perfectly-balanced model that is fully aware of all possible sources of bias and prejudices. 
It is reasonable to assume such a thing will not be achievable any time soon, mainly because, by definition, machine learning deals with ill-posed problems and incomplete data.
\textit{Counterfactual Fairness} carries excellent insight on how we can evaluate \textit{fairness} while also considering historical biases.
However, this metric requires a deeper analysis of the data in order to find the dependencies of features regarding both the sensitive attribute(s) and the outcomes.

\subsection{Metrics for Image Generation}
Image generation is the task of creating images given an input, be it a user-guided signal (including another image) or simply random noise.
Recent advances in the field have made it possible to create an image of almost anything given a simple description~\citep{ramesh2022hierarchical}.
Generative models are trained on colossal amounts of unbalanced data, which carry historical and societal biases. 
It is possible to measure whether a generative model is creating a similar number of samples for different groups by using a classifier on a large set of generated samples, verifying which group they belong to and then comparing these proportions.
\citet{Choi2020} used this strategy with a binary gender classifier to evaluate their face generation model trained on the \textit{CelebA} dataset, which has a higher proportion of females.
\citet{cho2022dall-eval} made a more comprehensive evaluation concerning gender and race using CLIP~\citep{radford2021learning} and human annotation.
That analysis used images generated through professional and political prompts to check whether the model correlated the protected attributes with specific roles. 
Overall, this approach requires a classifier with good performance in the respective groups so manual labeling is not necessary.

\subsection{Metrics for Language Modeling}
Language modeling (LM) is a base task for most NLP models.
By predicting the next token in a sentence, neural networks can learn to manipulate language, discovering multiple meanings of words that vary contextually.
A common approach is to train an LM and later adapt it to a downstream task, following the idea of first teaching the model about the nature of the language and then fine-tuning its knowledge to a specific task.
Instead of training a new model, it is possible to use only the generated embeddings (word or context-level) that encapsulate the semantics in a dense vector and then employ it in a different processing pipeline.

Many metrics use the LM output to estimate biases via the probabilities.
Another alternative is to use the dense vectorized representation to measure bias, which can be used both with contextual and word embeddings.
Besides those alternatives, NLP tasks can be used to measure bias when using specific evaluation  datasets.
Metrics tailor-made for LMs are described next.

\textbf{Direct Bias (DB)}~\citep{bolukbasi2016man} is a specific gender-bias metric which defines bias as a projection onto a gender subspace.
To measure DB, we first need to identify $N$ words that are gender-neutral.
Given those words and the gender direction learned $g$, we define the DB of a model as:
\begin{equation}
    DB = \frac{1}{|N|} \sum_{w \in N}|\cos (\vec{w}, g)|^c
\end{equation}
where $\vec{w}$ is the embedding vector of $w$ and $c$ is a user-defined parameter that determines how strict DB will be.
With $c = 0$, $DB=0$ only when there is no overlap of $\vec{w}$ and $g$.
With $c = 1$ we have a more gradual bias measurement albeit with a small error margin.  

\textbf{Word Embedding Association Test (WEAT)}~\citep{weat}  measures bias through the permutation of two sets of target words $X$ and $Y$ (\eg male-dominated professions like \textit{engineers} and female-dominated professions like \textit{nurses}) and two sets of attribute words $A$ and $B$ (\eg \{\textit{man},\textit{ male},...\} and \{\textit{woman}, \textit{female},...\}).
In a scenario without biases, there would be no difference between the relative similarity of the two sets of target words and the two sets of attribute words.
WEAT measures this difference in similarity to determine whether a given word is biased.

\textbf{Co-occurrence Metrics}~\citep{bordia2019wordlevel} uses word co-occurrence for measuring bias in generated text.
In a model-generated corpora, we can analyze the number of times that each word appear next to specific terms, \ie how often a language model will connect professions, sentiments and areas of knowledge with protected groups.

\textbf{Sentence Embedding Association Test (SEAT)}~\citep{seat} comes as the natural adaptation of WEAT~\citep{weat} but for contextualized word embeddings. 
Since WEAT only tested associations among word embeddings, it lost utility for recent models that are contextual-based (\eg based on Transformers).
The authors adapt it to make use of sentence templates, \eg \textit{\say{[He/She] is a [MASK]}}.
The models then generate contextual embeddings with these templates and the cosine similarity is computed between two sets of attributes.
Several other WEAT and SEAT variations have been proposed since, though with a similar usage principle.

\textbf{Discovery of correlations (DisCo)}~\citep{webseter2020measuring} uses a template with two slots, \eg \textit{\say{X likes to [MASK]}}, where \textit{X} is a word based on a specific set of words planned to trigger possible biases, and the \textit{[MASK]} token is replaced with the model prediction.
It compares the predictions for the words used in the template to compute biases. 
The final result is the average of different predictions for the sets of words.

\textbf{Log probability bias score}~\citep{kurita2019measuring} uses the prior probability of the model when predicting a specific \textit{[MASK]} token to normalize the resulting probability of a word that appears in a given context.
By doing so one can surpass limitations from metrics based on pre-trained models that do not consider the prior probability of a given word when generating the recommendation.
A limitation of this method is that it only works for models trained with masked language modeling, where the priors can be extracted by masking more than one token in the sentence, \eg~\textit{\say{The [MASK] is a [MASK]}}, and analyzing them individually.

\textbf{Context association test (CAT)}~\citep{stereoset} is a metric proposed in conjunction with the StereoSet dataset.
That dataset comprises sentences to be completed (model has to fill a blank token or select a continuation for the sentence).
The completion option, for all cases, contains stereotype, anti-stereotype, and meaningless options.
The objective of the evaluation is to measure how many times a model would choose a meaningful sentence over a meaningless one and how many times it would choose a stereotyped option instead of an anti-stereotype.

\textbf{CrowS-Pairs}~\citep{crows} is a \textit{pseudo-log-likelihood} metric based on the perplexity measure of all tokens conditioned on the stereotypical tokens.
Similar to the previous one, this metric also comes with a dataset.
The templates for its usage follow a similar approach to StereoSet, with stereotyped and anti-stereotyped versions.

\textbf{All Unmasked Likelihood (AUL)}~\citep{aul} is an extension of CrowS-Pairs.
It leverages not only the masked tokens in the sentence to measure biases but also all unmasked tokens and multiple correct predictions.
\citet{aul} also proposes AULA, a variation that takes into account the attention weights.

\subsubsection{Critical Analysis of the Language Modeling Metrics}
LMs are assets of high relevance in NLP, especially after the adoption of pre-trained LMs as the basis for adaptation in downstream tasks.
This \textit{reuse approach} highlights the need of a consistent and robust fairness evaluation due to its high-spread potential.
Metrics that rely on model probabilities are easier to interpret, especially when compared with the embedding-based metrics.
The latter are divided into two categories: word embeddings and contextual embeddings, which share quite a few similarities.
However, the distinctions become stronger as we look into metrics such as WEAT and SEAT, where we need to create specific scenarios to measure biases, generating a considerable level of human interference when evaluating the final embeddings.
The metrics presented here use either probabilities or embeddings and can be called intrinsic metrics.
However, they are not the only way to evaluate LMs.
We can evaluate fairness through downstream tasks using task-specific metrics. 
Neural machine translation, coreference resolution, and language generation are examples where we can use specific datasets and their respective evaluation protocols to analyze fairness levels.
Given the number of toxic terms in a text generated by a model, the quality of translations or the correlation of terms with protected attributes in complex and sensitive contexts may give a clearer picture of real-world problems that may occur.

\subsection{Task-Agnostic Metrics}
The following metrics do not rely on any specific task, \ie they are of general purpose and can be easily computed when evaluating trained neural networks.

\textbf{Bias Amplification}~\citep{zhao2017men} evaluates how much bias the model amplifies in a given dataset. 
For that, it measures the bias score $b(\hat{y}=c_k, z)$ as:

\begin{equation}
    b(\hat{y}=c_k, z) = \frac{c(\hat{y}=c_k, z)}{\sum_{z^{\prime} \in Z} c\left(\hat{y}=c_k, z^{\prime}\right)},
\end{equation}
where $c(\hat{y}=c_k, z)$ is the amount of times the model outputs $\hat{y}=c_k$ taking into account protected attribute $z$, and $Z$ is the set of protected attributes. 

The premise is that the evaluation set is identically distributed to the training set, and therefore if $\hat{y}=c_k$ positively correlates with $z$, and if $\tilde{b}(\hat{y}=c_k, z)$ (evaluation set) is larger than $b^*(\hat{y}=c_k, z)$ (training set), one can assume that the model has amplified that specific bias. 

Assume we are measuring biases in a VQA application, and that the bias scores measured in a specific model are $b^*(\hat{y}=\text{cooking},z=\text{woman}) = .66$ and $\tilde{b}(\hat{y}=\text{cooking},z=\text{woman}) = .84$, respectively, when asked \textit{What is the person doing?}.
In that case, one can assume that the model amplified the bias of \textit{cooking} toward \textit{woman}.
The authors do this for each class in order to obtain the mean bias amplification, defined as:

\begin{equation}
    \frac{1}{|K|} \sum_{z} \sum_{k \in\left\{k \in K\left|b^{*}(\hat{y}=c_k, z)>1 /\right| \mid Z \|\right\}} \tilde{b}(\hat{y}=c_k, z)-b^{*}(\hat{y}=c_k, z),
\end{equation}

However, the premise that the bias distribution will be the same from training to evaluation/test set might not hold and could misrepresent the bias-resilient capability of the model. 
We observed only one other work~\citep{hirota2022quantifying} that makes use of this metric outside the scope of the VQA(-CP) datasets.

\textbf{KL-Divergence}
The Kullback-Leibler divergence score quantifies how much a given probability distribution differs from another. 
The KL-divergence between distributions $P$ and $Q$, $KL(P \parallel Q)$ is calculated as:

\begin{equation}
    KL(P \parallel Q)=\sum_{i=1}^{N} P\left(x_{i}\right) \cdot \log \frac{P\left(x_{i}\right)}{Q\left(x_{i}\right)}.
\end{equation}

The divergence is an optimal metric to measure how different protected attributes diverge in a task where the feature should not correlate to the problem, \eg a \textit{curriculum vitae} system should not take gender into consideration~\citep{pena2020bias}.
Hence, if the distribution between genders significantly diverges, we can conclude that the model is unfair towards a gender.
Nevertheless, KL-Divergence is not a distance metric between two distributions since it is not symmetric. 
Therefore, $KL(P \parallel Q) \neq KL(Q \parallel P)$, requiring coupling KL-Divergence with another metric for bias mitigation measurement.

\section{Debiasing Methods for Fairness-Aware Neural Networks}
\label{sec:debias-methods}

\begin{figure}[!tp]
    \centering
    \includegraphics[scale=0.9]{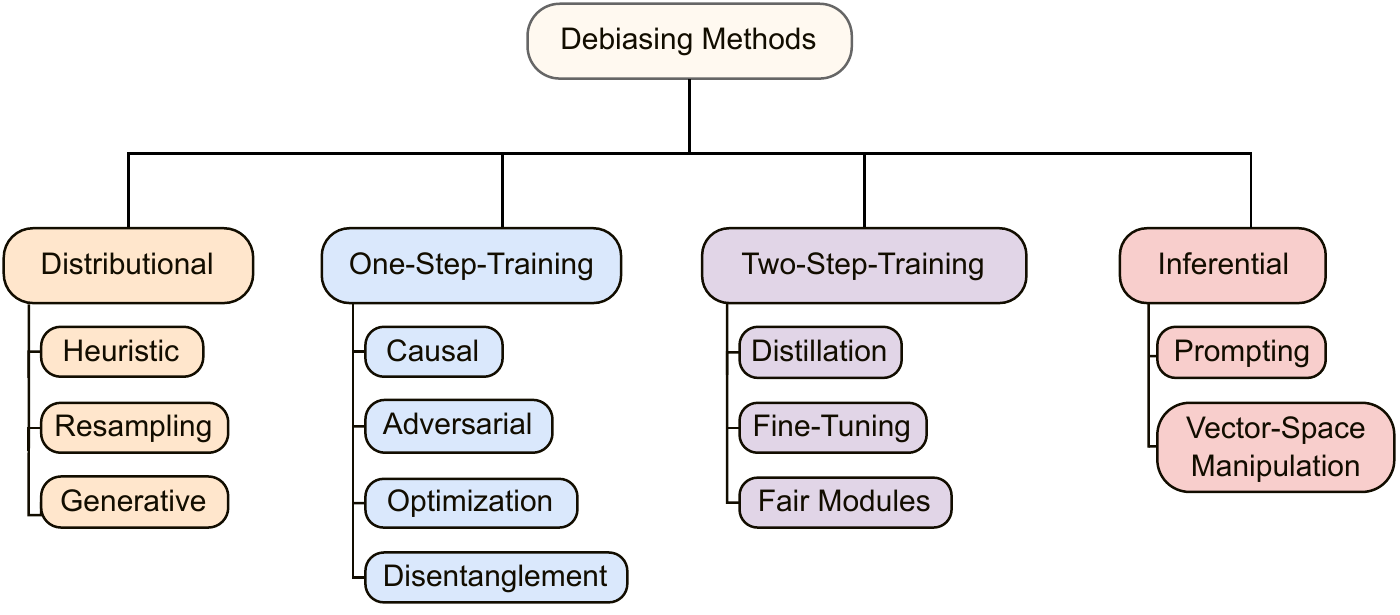}
    \caption{Taxonomy for organizing the literature on debiasing methods for neural networks in vision and language research.}
    \label{fig:taxonomy}
\end{figure}

This section organizes and discusses the existing literature for debiasing neural-network models in vision and language research. 
While the existing literature reviews divide the debiasing approaches into three categories, namely pre-processing, in-processing, and post-processing~\citep{mehrab2019survey,caton2020fairness}, we understand that these categories are insufficient to organize all existing methods reviewed in this survey in a precise fashion. 
Therefore, we propose a new taxonomy that properly categorizes all debiasing methods, and we present it in Figure~\ref{fig:taxonomy}. 

We call \textit{distributional} all strategies that modify the dataset distribution \textit{prior to training}.
That includes sampling strategies that increase the amount of data examples artificially. 
In addition, we divide methods that focus on optimization via training into two categories: 
\begin{enumerate}[label=(\roman*)]
    \item \textit{One-Step-Training}, which includes \textit{fair} models that are generated for a particular task via a single optimization procedure;
    \item \textit{Two-Step-Training}, where a new training phase has to be performed to fix an existing biased model, \ie making it \textit{fairer}.
\end{enumerate}
Finally, we call \textit{inferential} those strategies that address the problem of fairness based on the model outputs, \ie that discover and remove social biases without requiring further weight optimization or dataset manipulation.

Our taxonomy is more precise than previously-proposed ones in that it better differentiates methods that remove biases during the training process of the downstream task (one-step-training) from those that optimize a pre-trained model.
Note that when we say \textit{pre-trained}, we mean trained in the final (goal) downstream task, not pre-trained in large general datasets (say Imagenet~\citep{olga2015imagenet}).
Hence, the difference between one-step-training and two-step-training is whether there is a previous biased model that works for the \textit{task of interest} or not.
Methods that fall in the \textit{one-step-training} category may very well have been pre-trained on general-purpose datasets.
This updated taxonomy reflects the fast advances we are witnessing in both industry and research communities with the adoption of the so-called \textit{Foundation Models}~\citep{bommasani2021opportunities}.

Finally, we should mention that the categories of the taxonomy are not mutually exclusive. 
Let us assume a given method uses an adversarial strategy coupled with a regularization term for debiasing a model pre-trained on Imagenet.
We categorize it as belonging to both \textit{adversarial} and \textit{optimization} categories within \textit{one-step-training}. 

Table~\ref{tab:debiasing-methods} organizes all debiasing methods surveyed in this paper according to the proposed taxonomy and also the respective domain (vision, language, or multimodal). 

\begin{table}[!bp]
    \centering
    \scriptsize
    \caption{Debiasing methods organized according to the proposed taxonomy.}
    \label{tab:debiasing-methods}

    \begin{tabular*}{\textwidth}{@{\extracolsep{\fill}}ccccc}
    \toprule
    Category                            & Sub-category                      & Vision                                                                                                                                                & Language & Multimodal \\ \midrule
    \multirow{3}{*}{Distributional}     & Heuristic                         & \cite{derman2021dataset}                                                                                                                              & \cite{cda,tga}   & ---          \\
                                        & Generative                 & \cite{Ramaswamy2021,gambs2022fairmapping,ngxande2020bias,yucer2020exploring,Chen2021}                                & \cite{qian2022perturbation}   & ---          \\
                                        & Resampling                        & \cite{cao2022fair,Li2019,tanjim20223908}                                                                                                                             & ---   & \cite{yan2020mitigating}          \\ \midrule
    \multirow{5}{*}{One-Step-Training}  & Adversarial                       & \cite{edwards2016censoring,Wang2018,li2020deep}                                                                                                       & \cite{Odbal2022422,gaci2022iterative,wu2021fairness,rekabsaz2021societal,fleisig2022mitigating,maheshwari2022fairnlp,zerveas2022mitigating} & \cite{yan2020mitigating,berg2022prompt} \\
                                        & Causal Approaches                 & \cite{kang2022fair,dash2022causalperspective,kim2021counterfactual}                                                                                   & \cite{gupta2022distilled}   & \cite{yang2021causal}          \\
                                        & Disentanglement                   & \cite{Park2021,Creager2019,Xu2020,Tartaglione2021,kim2021counterfactual}                                                                              & \cite{du2022fairdisco}   & ---          \\
                                        \cmidrule{2-5}
                                        & \multirow{2}{*}{Optimization}     & \multirow{2}{*}{\cite{martinez2020minimax,tanjim20223908,wang2020mitigating,Agarwal2022,hendricks2018women,amini2019uncovering,gronowski2022utility}} & \cite{Li2022debiasingneural,du2022fairdisco,shen2021contrastive,zerveas2022mitigating,bordia2019wordlevel,kaneko2021debiasing,webseter2020measuring,chen2022unsup} & \multirow{2}{*}{\cite{wang2021gender,kim2022information,zhao2017men}} \\
                                        &                                   &                                                                                                                                                       & \cite{patil2022decorrelation, serna2022sensitiveloss,li2020deep,Kim2019,kenfack2022repfair,ma2021conditional,mcgovern2021source,qian2019reducinggender} & \\ \midrule

    \multirow{3}{*}{Two-Step-Training}  & Distillation              & \cite{Mazumder2022,li2021learning,Jung2021}  & \cite{gupta2022distilled}   & ---          \\
                                        & Fair-Modules              & \cite{karakas2022fairstyle,li2021learning}  & \cite{geep,wu2021understanding,fairfil,adele,rajabi2022looking}   & \cite{yan2020mitigating,park2020fair,tang2021mitigating}  \\
                                        & Fine-Tuning               & ---  & \cite{gira2022debiasing,faal2022reward,liu2021reinforcement,wu2022fairprune}   & \cite{berg2022prompt}          \\ \midrule
    \multirow{3}{*}{Inferential}        & Prompting                 & ---  & \cite{wallace2019universal,sheng2020towards,gehman2020real,sharma2022sensitivetranslation,schick2021self}   & \cite{mishkin2022risks}  \\
                                        \cmidrule{2-5}
                                        & \multirow{2}{*}{Vector-Space Manipulation} & \multirow{2}{*}{\cite{salvador2022faircal}}  & \cite{liang2020towards,liang2021towards,bolukbasi2016man,Du2020mdr,aekula2021double,dev2021oscar,Vargas2020exploring,bolukbasi2016quantifying,kaneko2019gender}   & \multirow{2}{*}{\cite{wang2021gender}}  \\
                                        & & & \cite{Gyamfi2020deb2viz,kumar2020nurse,yang2020causal,dev2019attenuating,Subramanian2021evaluating,kumar2021hiperbolic,kaneko2021dictionary,lauscher2020general} &    \\ 
    \bottomrule  
    \end{tabular*} 
\end{table}

\subsection{Distributional}
Bias-mitigation methods of the category \textit{distributional} target at changing the data distribution to better represent socially-vulnerable groups or to change data objects of the dataset to remove unwanted biases. 
They rely on creating a modified dataset to improve fairness or applying systems and rules to remove or compensate underrepresented groups within the data. 
The rationale is that the neural network will be trained using a more representative data distribution, thus leading to a fairer model.

Distributional mitigation methods are stratified into 3 groups according to the changes made to the data: 
\begin{enumerate}[label=(\roman*)]
    \item \textbf{Heuristic} methods modify objects of the dataset according to predefined algorithmic rules;
    \item \textbf{Generative} methods aim to create or modify data objects using generative models;
    \item \textbf{Resampling} methods are based on under or over-sampling the dataset to mitigate the under-representation of individuals in protected groups.
\end{enumerate}

\subsubsection{Heuristic}
Unbalanced training sets may lead to models that reproduce certain disparities present within the data.
For that, one might change the dataset by modifying, adding, or removing objects, somehow making it represent all protected groups/individuals equally. 
When making these modifications, one should consider that not all applications follow the same distributions, and hence are prone to the same set of rules.
For instance, new sentences must follow pre-existing grammatical, lexical, and syntactic rules when one is modifying NLP datasets.

Debiasing through heuristic manipulation is a viable strategy for NLP tasks.
Examples include using dictionaries or semantic and syntactic rules to replace words or add informational labels/tokens~\citep{cda,tga}. 
More than just replacing words, those strategies seek to create new information, not only by changing one or two words but by adapting the entire surrounding information, avoiding the creation of nonsensical sentences.
For computer vision, a possible heuristic approach is to apply pre-determined transformations in images to augment the dataset. 
Some methods can leverage this process to augment only instances underrepresented in the dataset~\cite{derman2021dataset}.

While heuristic methods to change data distribution may be an interesting alternative for bias mitigation, these changes are only practical in scenarios where the application domain comprises explicit and well-known rules. 
One example is language modeling, where templates of sentences can be used to expand the original data. 
In other domains, however, heuristics do not scale and fail when the rules are not easy to define manually.

\begin{figure}[!bp]
    \centering
    \includegraphics[scale=0.35]{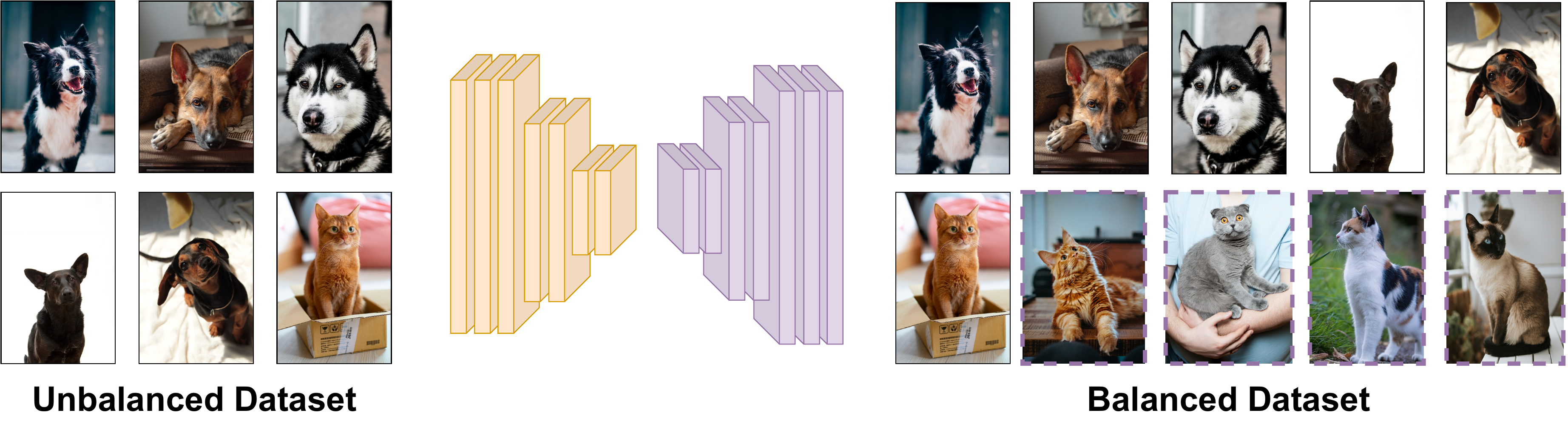}
    \caption{General overview of generative methods for debiasing.}
    \label{fig:generative}
\end{figure}

\subsubsection{Generative}
In domains like computer vision, where no explicit set of constraints is specified to delimit data distribution, more complex approaches may be necessary to adjust the level of fairness in a dataset.
Finding a simple strategy or rule that can be applied to all instances is hardly possible, requiring the creation of deep networks capable of doing such modifications~\citep{Ramaswamy2021}.

Generative Adversarial Networks (GANs) can be an option when looking for ways to increase a dataset with synthetic data~\cite{Ramaswamy2021,gambs2022fairmapping,ngxande2020bias,yucer2020exploring}, since they can create high-quality new images when properly trained, balancing the dataset with regard to its potential misrepresentation and allowing the training of a new model over both original and synthetic data. 
Figure~\ref{fig:generative} gives an overview of this strategy of augmenting (or balancing) a dataset with a generative neural network.

In the language context, \citet{qian2022perturbation} propose a sequence-to-sequence model that generates perturbations regarding the protected variables in dataset instances. 

Systematic biases exist in many datasets, which is also the case with face datasets.
As annotators usually consider female faces as being happy more often than men's, \citet{Chen2021} propose using Action Units (AU) that aim to measure facial expressions to address this problem objectively.
The proposed framework does not need to modify labels, like other methodologies.
Using AUs, the model classifies two similar AU samples as similar people besides their gender.
To penalize for unfairness, they utilize the triplet loss function as a regularizer.
Note that this method can be applied to other datasets (not only facial expressions), and one just needs to have other objective measures, like body key points.
That work was the first to show the systematic effect of annotations in datasets for computer vision, and it is even more visible in in-the-wild datasets.

\subsubsection{Resampling}
Resampling debiasing methods only modify the data distribution by rearranging the existing dataset objects.
While possibly lowering the number of objects available for training, no artificial data is generated.
The most common resampling approaches balance the distribution of the training dataset according to a specific attribute such as gender or race~\cite{yan2020mitigating} while others focus on discovering how to resample the original dataset to create a fairer distribution that is better suited to represent all protected groups in the data~\cite{Li2019,tanjim20223908}.

A more sophisticated strategy is to use more than one dataset to build a more representative one.
This process can use different sampling approaches to select the best configuration of available objects to better represent the diversity of protected attributes, thus providing a fairer dataset for training~\cite{cao2022fair}.

\subsection{One-Step-Training}
Although manipulating the data is a straightforward strategy to have more diverse or balanced data, it is often not enough to produce fair neural models.
Some challenges that remain unsolved by distributional approaches are: 
i)~(deep) neural networks are data hungry, which means undersampling strategies could reduce the data up to the point training becomes unfeasible; 
ii)~even with data that perfectly represents the population distribution, undesirable characteristics such as stereotypes and prejudice that are present in society may arise~\citep{WangT2019}. 
For solving those problems, one may need to resort to additional strategies that happen either during training or during inference, which are the focus of this section.

We call \textit{one-step learning} debiasing methods that act during the main training procedure.
These methods are further divided into four distinct groups according to the debiasing strategy that is used:
\begin{enumerate}[label=(\roman*)]
    \item \textbf{Adversarial} methods make use of the adversarial framework or of adversarial examples to teach the model not to resort to undesired biases;
    \item \textbf{Causal} methods use causal graphs and counterfactual examples to teach the model which relationships are relevant within the data;
    \item \textbf{Disentanglement} methods separate the features in the latent space to manipulate them independently;
    \item \textbf{Optimization} methods include loss function adaptions, addition of regularization terms, and other modifications for improving weight optimization.
\end{enumerate}

\subsubsection{Adversarial}
\begin{figure}[!tp]
    \centering
    \includegraphics[width=\columnwidth]{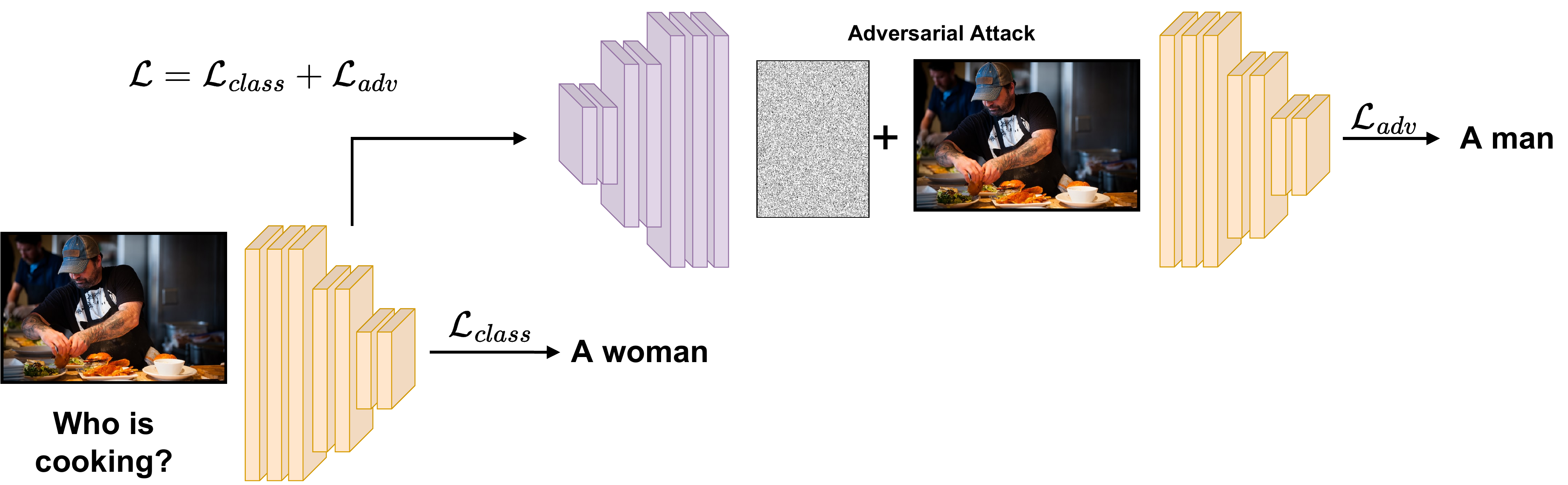}
    \caption{General overview of adversarial methods for debiasing.}
    \label{fig:adversarial}
\end{figure}

It is a known fact that adversarial examples can deceive deep learning models. 
Neural networks may be fooled by intended perturbed images that do not contain human-perceivable changes~\citep{goodfellow2015explaining}. 
Notwithstanding, adversarial examples can be included in the training dataset to create more robust models. 
That setup uses two models: one is trained to solve the task objective, whereas the other focuses on creating adversarial examples that try to confound the first model~\cite{goodfellow2014generative}. 
The objective is to teach the main network not to use the protected attribute to do the task~\citep{xu2019achieving,zhang2018mitigating, Odbal2022422,gaci2022iterative,wu2021fairness,li2020deep}. 
To force the model not to rely on protected attributes, it is possible to erase or mask them from the data source~\citep{Wang2018}, create new data, or even attack the deep representation generated by the model~\cite{edwards2016censoring,fleisig2022mitigating,maheshwari2022fairnlp}, which is less interpretable. Figure~\ref{fig:adversarial} depicts this general idea.

For the information retrieval domain, adversarial training is being used to create embeddings that are bad predictors for the protected attribute, but that are capable of accurately predicting the target variable~\cite{rekabsaz2021societal, zerveas2022mitigating,yan2020mitigating}.

Note that all adversarial methods rely on annotations of protected variables or groups in the dataset.
They allow the models to focus on legitimate attributes while ignoring the protected ones.
However, the need for annotation is a downside of these approaches.
Fortunately, the annotations are not needed for inference time.

\subsubsection{Causal}
We can leverage any knowledge on the \textit{causal-effect} relationship between protected attributes and outcomes to try and fix model unfairness. 
For identifying these relationships one can make use of causal graphs or counterfactual samples during training. 
Causal approaches are a popular choice to mitigate general biases, such as language priors in VQA~\cite{kolling2022efficient,chen2020counterfactual,abbasnejad2020counterfactual,niu2021counterfactual}, and
we explore how they can be adapted to solve fairness issues. 

By using causal graphs we can force the model to learn valuable relationships between features, intermediate representations, and predictions. 
\citet{yang2021causal} propose a specific attention module in the network that is responsible for discovering such associations, and then use them for bias mitigation. 
Another benefit of building a causal graph is that it can later be used to explain predictions, an important artefact for \textit{explainable AI}.

Creating counterfactual samples, on the other hand, allows the model to train over examples that are harder to predict, since they may not occur in the training distribution, improving the robustness of the model towards specific protected groups.
We can employ linguistic rules to create new examples during training, forcing the model to learn with one group and its opposite (\eg male and female genders)~\cite{gupta2022distilled}. 
Other strategy is to disentangle visual representations in the latent space and use isolated features to create the counterfactual samples~\cite{kim2021counterfactual}.
Additional work uses an auxiliary model to discover the causal graph and use discovered relations to create counterfactual samples using the protected attributes~\cite{kang2022fair,dash2022causalperspective}.

Generating counterfactual samples is a good option to ensure that the model will see a larger number of distinct examples.
However, such approaches may lead to creating extreme counterfactuals that do not represent valid instances from the dataset or the distribution they represent~\cite{king2006dangers}.

\subsubsection{Disentanglement}
During training, neural models create latent representations that represent automatically-learned features extracted from the data objects.
The difficulty of learning a task and the robustness and generalizability of a model are directly correlated with the quality of the respective learned latent representations.
One way to increase such quality is via disentanglement.
A \textit{disentangled} representation is a representation where each learned factor corresponds to a single factor of variation in the data and is invariant to other factors of variation~\cite{bengio2013representation}.
Disentangled representations offer many advantages, such as boosting predictive performance~\cite{locatello2019challenging}, increasing interpretability~\cite{higgins2017betavae}, and improving fairness~\cite{locatello2019fairness}.

Learning disentangled representations means to break down features into new dimensions where information is split into independent spaces. 
This approach can be used to separate the underlying structure of some objects in their various parts in a more representative and interpretable way~\citep{Higgins2018}.
For instance, the shape of an object and its position in the image can be independently broken down. 
That isolation of features allows us to operate in specific details of the input data rather than modifying it entirely. 
StyleGAN~\cite{karras2019style} is an example of that, where it is possible to modify only the hair attributes of someone's face without further changes to the image. 

\begin{figure}[!tp]
    \centering
    \includegraphics[width=\columnwidth]{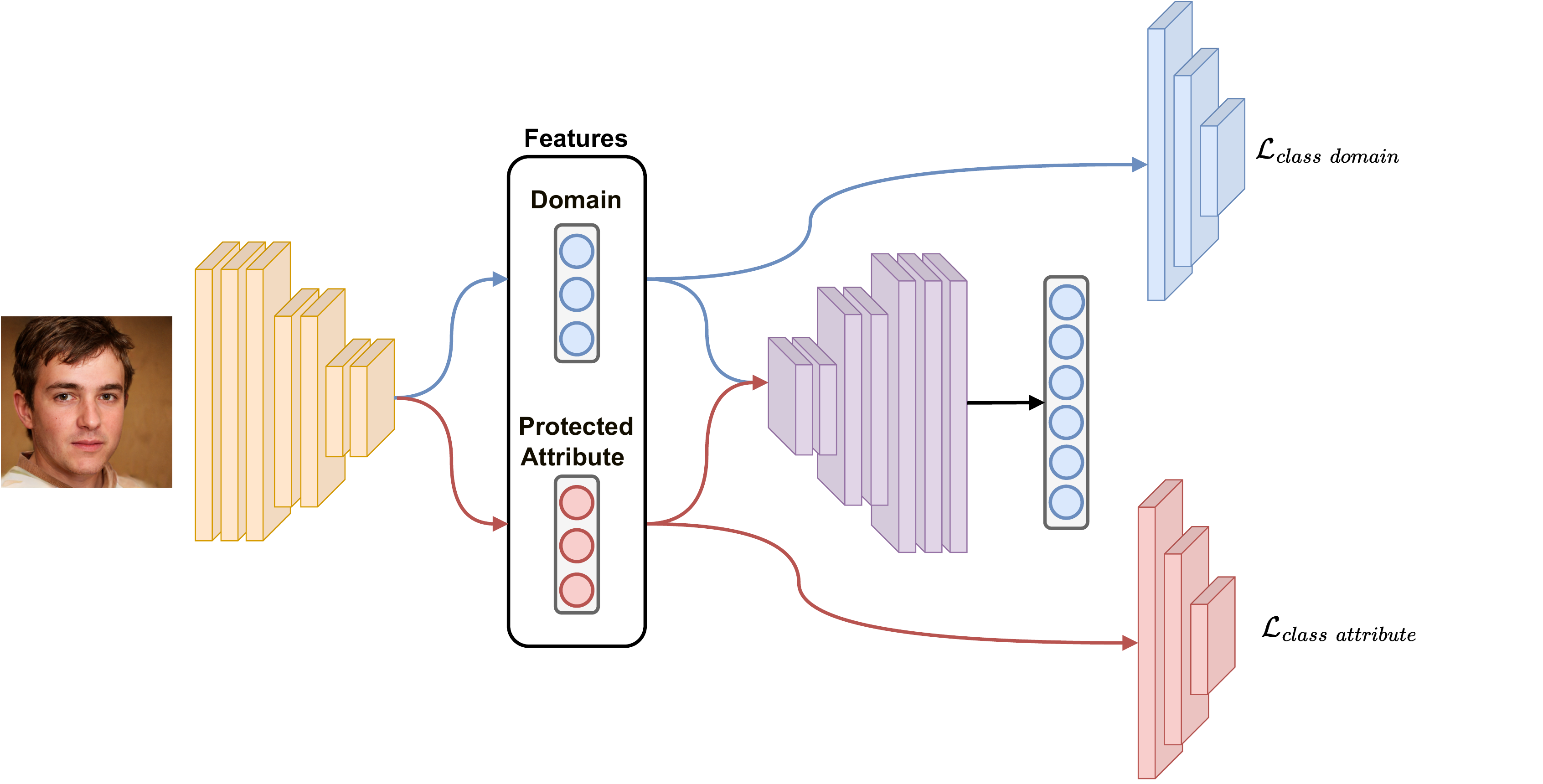}
    \caption{General overview of the disentanglement approach for debiasing.}
    \label{fig:disentanglement}
\end{figure}

That idea of splitting different data dimensions can also be explored for debiasing during training, inducing a model to learn only features that are invariant to the protected attributes. Figure~\ref{fig:disentanglement} illustrates such an approach.

By decomposing the representation into different spaces and splitting the target attribute dimension from the protected attribute dimension, a model may learn not to correlate sensitive information to the real objective task, prioritizing only relevant and essential information. 

This approach can be implemented as a new model or module responsible for the disentanglement~\cite{Creager2019} or as a regularization term that uses the labels to guide the division of features~\cite{Tartaglione2021}.

\citet{Xu2020} show that disentangled latent representations may achieve superior performance in terms of demographic bias than simply adding the protected attributes in the classification layer in a typical fairness-through-awareness approach. 
In the study of \citet{Park2021}, a variational autoencoder (VAE) is employed to create a representation that disentangles the latent space into three distinct subspaces:
\begin{enumerate}[label=(\roman*)]
    \item The first one has information of the target variable and no protected attribute information;
    \item The second contains both target and protected attribute information;
    \item The third contains just protected attribute information.
\end{enumerate}
  
Images often carry both protected and legitimate attributes that are hard to separate.
Disentanglement learning is an alternative to achieving fairness through unawareness in computer vision
\citet{du2022fairdisco} propose to disentangle the latent space and to use only the subspace that spans the legitimate factor to perform the classification task. 
\citet{kim2021counterfactual} implement the concept of counterfactual fairness while requiring the counterfactual samples to generate disentangled representations of the attributes. 

Disentanglement approaches are a good option for separating protected and non-protected features, making it possible for the model to consider only one group of attributes when doing a specific task.
Unfortunately, to perform a task such as classification, the training phase must be divided into at least two steps: the first one to learn the disentangled representations and the second to learn the target task effectively.

\subsubsection{Optimization}
Optimizing a loss function is the training strategy of every neural network. 
Together with an optimization rule, they dictate how the weights must be learned for a specific task. 
By adding or modifying terms in a loss function, one can drastically change the solutions generated by a neural network. 
We categorize a method as being an optimization approach if it proposes changes in the optimization procedure during training towards increasing fairness, forcing it to follow a desired output distribution without modifying the input data~\cite{Jain2021}.

Several studies propose new loss functions for penalizing or ensuring specific behaviors. 
Often these new loss terms are summed or used together with the objective loss (\eg Cross-Entropy), ensuring that the model will learn the main task while also being subject to fairness constraints. 
One can use this strategy to penalize models that wrongfully predict the protected attribute~\cite{hendricks2018women}. 
It is also possible to re-calibrate the loss function according to the value of the protected attribute~\cite{amini2019uncovering,tanjim20223908} or to optimize for fairer data representation~\cite{kim2022information,gronowski2022utility,zerveas2022mitigating,patil2022decorrelation,serna2022sensitiveloss,bordia2019wordlevel}.

With textual data, changes in loss function usually aim to remove bias from embedding representations. 
It is possible to manipulate the multi-dimensional space in which the embeddings are located to decouple them from non-desired sub-spaces~\cite{kaneko2021debiasing,webseter2020measuring, chen2022unsup,qian2019reducinggender,mcgovern2021source}.

Aside from proposing novel loss functions, it is possible to employ an algorithm to inject constraints during model training to ensure a fair distribution of predictions for the training data~\cite{zhao2017men,li2020deep}. 
Also, regularization terms can minimize the mutual information between feature embedding and bias~\cite{Kim2019}.
Other possibility is to apply a norm clip based on data groups, which improves diversity in data generated by GANs~\cite{kenfack2022repfair}.

Contrastive learning is also an optimization procedure capable of model debiasing. 
During training, the model faces examples from different classes, and it should classify them correctly while keeping examples from different classes apart in the embedding space.
We can use this technique to increase fairness during training by picking both positive and negative examples conditioned on the protected attribute~\cite{ma2021conditional,wang2021gender} or by enforcing that distinct protected groups be distant in the embedding space~\cite{shen2021contrastive}.

Optimizing weight distributions for particular examples in order to penalize easier ones is also a common strategy for debiasing~\cite{Li2019,Li2022debiasingneural,wang2020mitigating}. 
It is possible to reduce unfairness by maximizing ratios between losses in the reweighted dataset and the uncertainty in gold-standard labels. 
This strategy can come with a coefficient of variation as a data repair algorithm to curate fair data for a specific protected class and improve the actual positive rate for that class~\citep{Agarwal2022}. 
It has the advantage of working in both supervised and unsupervised learning.

Some studies aim to formulate the question of fairness as a multi-objective optimization problem.
The work of \citet{martinez2020minimax} defines each optimization objective as the protected group conditional risk.
In such formulation, the goal is to find solutions at the Pareto frontier.

With fairness criteria tied to the training objective of the network, two challenges arise: 
\begin{enumerate}[label=(\roman*)]
    \item There may be a demand for extra annotations in the data regarding the protected attributes to allow for computing novel losses;
    \item The optimization modifications can incur in a trade-off between fairness and accuracy, which we must be aware of.
    This is further discussed in Section~\ref{sec:challenges-future-directions}.
\end{enumerate}

\subsection{Two-Step-Training}
It is customary (and even considered a best practice) to not always train neural networks from scratch.
Instead, whenever possible, the weights of a model are set to the state of a previous training procedure, which may even have happened using a different dataset.
This typically accelerates training and often improves the performance and generalization capabilities of a model.
Since this practice is frequently adopted due to its benefits, applying debiasing methods during the initial stages of model training may be unfeasible or undesirable, especially since this may imply ``wasting'' computational resources that were used in the pre-training phase.
Thus, debiasing strategies that focus on adapting existing models become attractive to improve fairness.
We have separated two-step training methods into three distinct groups, according to the debiasing strategy used:
\begin{enumerate}[label=(\roman*)]
    \item \textbf{Distillation} includes methods that train a new model using the teacher-student approach;
    \item \textbf{Fair Modules} aggregates methods that add new modules to existing models to remove unfairness;
    \item \textbf{Fine-Tuning} lists approaches to retrain a model without architectural changes.
\end{enumerate}

\subsubsection{Distillation}
The core concept of distillation is to transfer knowledge between two models: the teacher and the student.
The student network is optimized to match the teacher model's predictions in addition to a task cost function.
The student and teacher models have similar architectures, but the student network has less capacity, \ie less optimizable parameters.
In this scenario, the distillation process creates a model with similar predictive capabilities while simultaneously reducing the amount of resources needed~\cite{sanh2019distilbert}.
One of the benefits of this type of approach is increasing the generalization capabilities of models without requiring more annotated data~\cite{hinton2015distilling}.

\begin{figure}[!tp]
    \centering
    \includegraphics[scale=0.35]{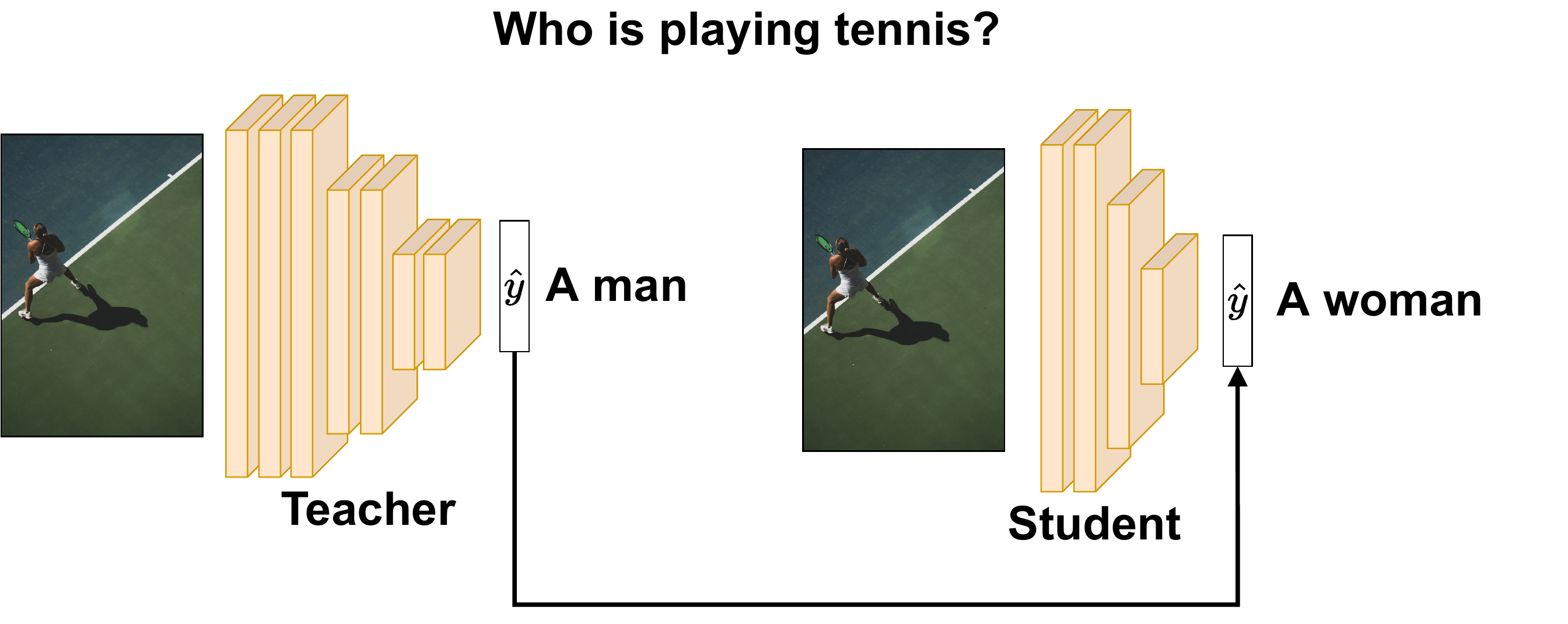}
    \caption{General overview of the distillation approach for debiasing.}
    \label{fig:distillation}
\end{figure}

In the scope of fair deep learning, one can use model distillation to produce fairer student models based on an unfair teacher.
To that end, most studies add a specific fair-related loss~\citep{Mazumder2022, gupta2022distilled, li2021learning} or regularization term~\cite{Jung2021} to the original distillation framework.
Intuitively, the student model learns to emulate the teacher model's knowledge while being simultaneously instructed to not rely on protected attributes present in the dataset.
Figure~\ref{fig:distillation} illustrates the core ideas of this type of approach.

Despite its benefits, the distillation strategy also presents some drawbacks.
For one, it is typically much more expensive than other available two-step training approaches.
In scenarios where the intended teacher model's weights or (log-)likelihood outputs are not available, \eg models behind APIs such as GPT-3~\cite{brown2020language}, it becomes impossible to train a student network.
Finally, the debiasing loss may conflict with the distillation objective, which may result in generalization problems~\cite{stanton2021does}.

\subsubsection{Fair Modules}
\begin{figure}[!tp]
    \centering
    \includegraphics[scale=0.5]{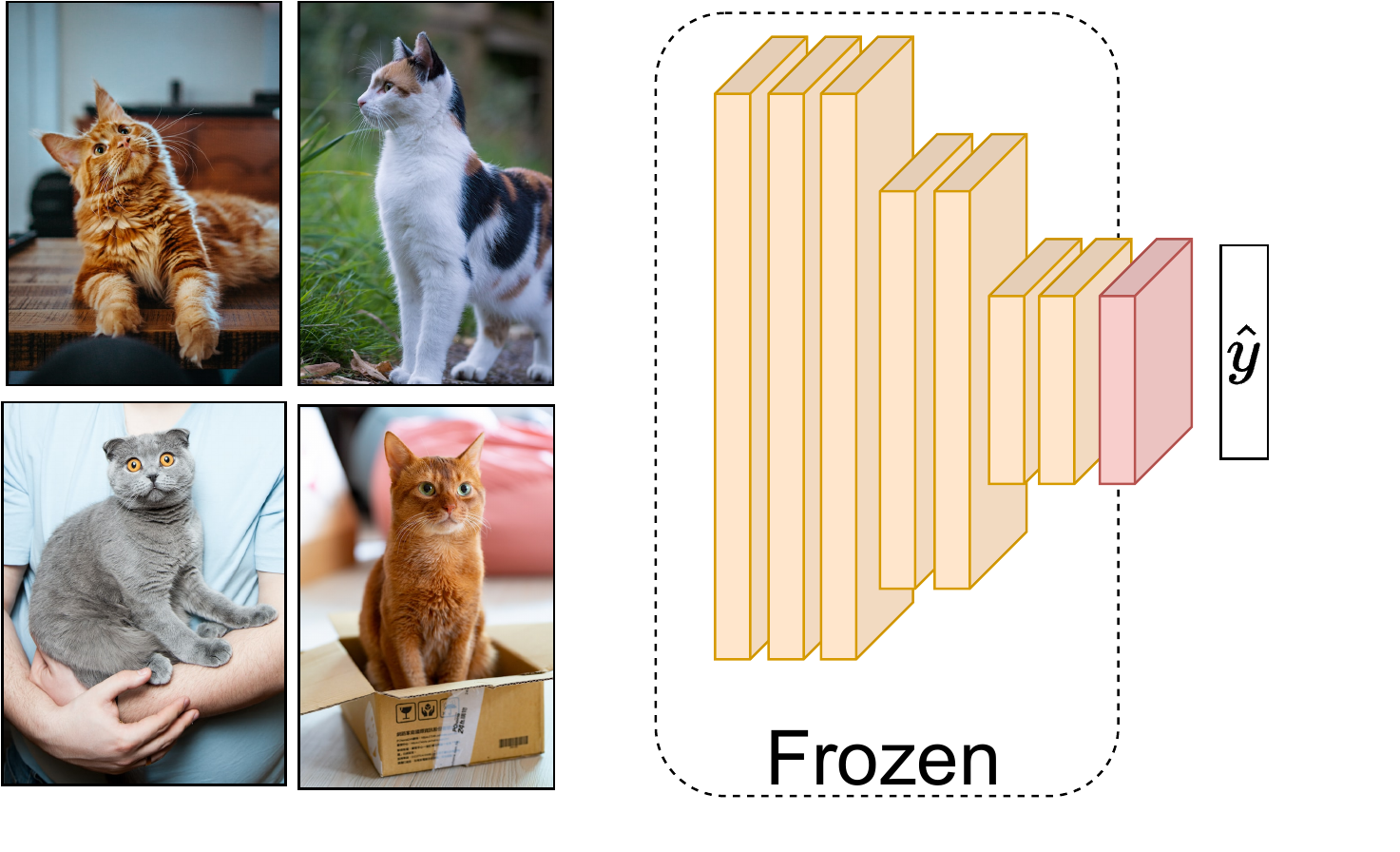}
    \caption{General overview of the fair-module strategy for debiasing.}
    \label{fig:fairmodules}
\end{figure}

Another useful group of debiasing techniques for pre-trained models involves adding new components, or modules, to an existing architecture, as depicted in Figure~\ref{fig:fairmodules}.
By combining the original architecture with a new group of layers, we can transfer the responsibility of learning fairness objectives to the new modules, and potentially leave the original weights untouched.

One may use specialized modules to create additional representations that can improve the quality of embeddings, thus preventing the model from learning \say{shortcuts} that lead to biases~\cite{wu2021understanding}.
Another potential approach is to include modules that detect or predict the presence of protected attributes, which is then used to modulate the overall answer~\cite{park2020fair, tang2021mitigating, karakas2022fairstyle,li2021learning}. 
Alternatively, some approaches add modules that are responsible for learning how to mitigate biases based on the inference of the original model~\cite{fairfil,adele,geep}.
Finally, \citet{rajabi2022looking} proposes to train an encoder-decoder module that processes inputs before the original classifier network to remove unwanted biases without changing the original model.
Since the new module removes potentially protected attributes, they tend to not affect the final model's prediction, effectively removing their influence.

One of the main advantages of this strategy is that one can freeze the original model weights and prevent model degradation and catastrophic forgetting, a phenomenon where previously-learned concepts and tasks are replaced by new information~\cite{geep}.
However, depending on the original model capacity, these architectural modifications may not work as well as techniques that adapt the original model weights.

\subsubsection{Fine-Tuning}
\label{sec:fine-tuning}
\begin{figure}[!tp]
    \centering
    \includegraphics[scale=0.4]{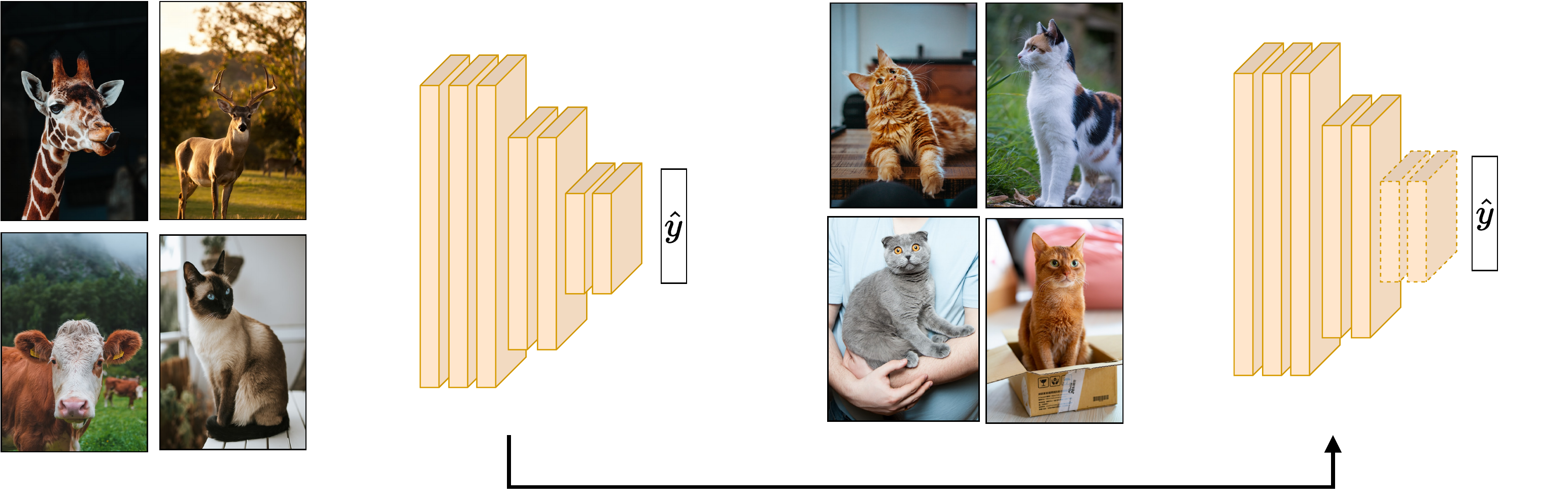}
    \caption{General overview of the fine-tuning strategy for debiasing.}
    \label{fig:finetuning}
\end{figure}

Fine-tuning refers to adjusting some or all of the weights of a model, that has already been trained, for a new combination of task, dataset, and domain.
Figure~\ref{fig:finetuning} shows the general framework of the fine-tuning strategy for debiasing neural models.
In the context of fairness, we can view fine-tuning as a specialization procedure.
That is, pre-training is responsible for teaching general concepts pertaining to a data modality (\eg language modeling for NLP), while fine-tuning forces the model to shift focus to fair concepts.

\citet{gira2022debiasing} has shown that fine-tuning a GPT-2 model changing as little as 1\% of the model's parameters using a curated dataset is enough to increase model fairness while retaining the knowledge acquired in the pre-training stage.
Reinforcement learning algorithms may also be used to guide weight optimization and provide a better alternative than a simple loss or dataset modifications~\cite{faal2022reward, liu2021reinforcement}.
\citet{wu2022fairprune} hypothesizes that some network parameters may be correlated with biased predictions, and thus consider model pruning to be a potential solution.
Paired with a specific dataset used as a guide, the authors prune a pre-trained model, removing weights responsible for such biased predictions while attempting to maintain a similar performance.

Differently from adding a new module to be responsible for debiasing, in fine-tuning approaches the driving factor is the dataset used to continue model training.
In situations where a sufficiently large and complex dataset is not available, fixing the problems of a pre-trained model using fine-tuning approaches may not be possible.
Additionally, debiasing via fine-tuning requires that the practitioner be aware of fairness problems that are already present in the pre-trained model (or if the model is affected by a specific bias of interest), which may imply a more thorough investigation via empirical analyses.

\subsection{Inferential}
Debiasing strategies that are applied during training or by changing the data distribution may be unfeasible in situations where the model is not available or when fine-tuning is a resource-intensive task.
Large models such as GPT-3~\cite{brown2020language} are not easily used by most practitioners due to hardware constraints.
For instance, to instantiate an OPT~\cite{zhang2022opt} model, that has around $130$ billion parameters, more than $300$GB of VRAM are required, which makes it impractical to fine-tune for most use cases.
Additionally, some large neural models are not open sourced and can only be accessed via commercial APIs, which makes unfeasible the application of the debiasing approaches presented so far.
Despite their generalization capabilities, such large models are also prone to accruing societal biases~\cite{parrots}.
Inferential methods are alternatives in such scenarios, because they intervene during inference time to make models fairer, leaving weights untouched.
There are two possible alternatives to this approach:
\begin{enumerate}[label=(\roman*)]
    \item \textbf{Prompting} prepends or alters the model input with specific triggers that stimulate a bias-free result;
    \item \textbf{Vector-Space Manipulation} manipulates the embedding space to remove undesired biases.
\end{enumerate}

\subsubsection{Prompting}
\label{sec:prompting}
When researching Foundation Models in NLP, \citet{brown2020language} discovered that this category of large-scale models can perform few-shot learning when receiving specific instructions in their prompts~\cite{brown2020language,zhang2022opt,black2022gptneox}.
In the context of FMs, prompts refer to the model input, and typically consist of tokens that are later converted to continuous representations (\eg an embedding layer).
This discovery started a separate field of study known as prompt engineering~\cite{reynolds2021prompt}.

Besides the simple approach of limiting and filtering a model's prompt (\eg by restricting the use of certain words)~\cite{mishkin2022risks}, recent studies have shown that language models are vulnerable when \textit{attacked} with specific tokens in their prompts, a procedure known as triggers~\cite{wallace2019universal}.
Triggers consist of tokens appended to the original prompt that lead to unexpected (or, in the case of fairness, desired) behaviors.
One can use such induced behaviors to mitigate (or provoke) certain biases, which make triggering methods attractive to promote fairness in large models~\cite{sheng2020towards}.
Recent studies~\cite{berg2022prompt, gehman2020real, sharma2022sensitivetranslation} seek to discover specific tokens (discrete or continuous) to help debiasing models.
\citet{schick2021self} propose an unusual form of debiasing through prompting by using the model itself to identify and mitigate its own biases. 
In that study, the authors prepended templates such as ``the following text discriminates against people because of their gender:'' to the prompt and identified that this addition manipulates the distribution of words to lower the probability of discriminatory outputs.

\begin{figure}[!tp]
    \centering
    \includegraphics[width=\columnwidth]{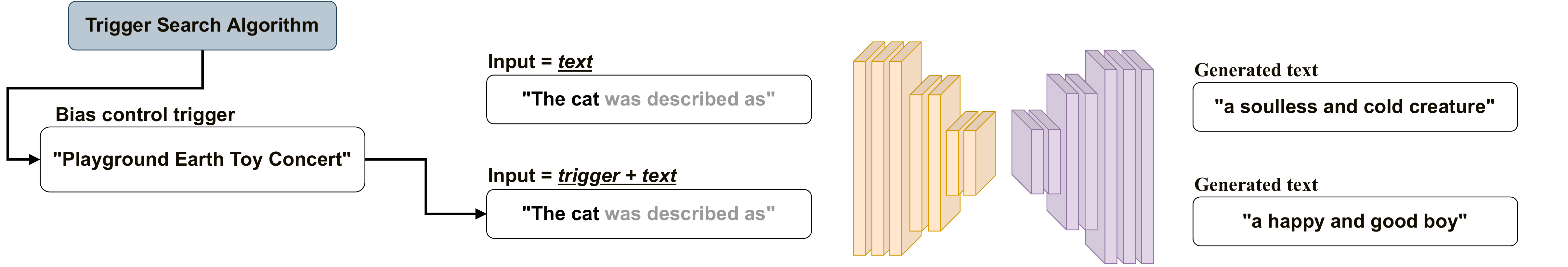}
    \caption{General overview of the prompting strategy for debiasing.}
    \label{fig:prompting}
\end{figure}

This is a versatile and attractive approach since, with the correct triggers, it is possible to mitigate or even induce specific biases, and potentially deal with multiple protected groups.
Figure~\ref{fig:prompting} shows the general idea of this category of methods.
A downside of prompting approaches is clearly exemplified in the context of NLP: triggers often add undesired contexts to sentence generation, since the model is conditioning its outputs on a token that may go against the initial context.

\subsubsection{Vector-Space Manipulation}
A common approach in NLP to learn the semantic meanings of words is to use word-embedding representations, which represent tokens as $d$-dimensional vectors~\cite{mikolov2013efficient, pennington2014glove}.
Compared to traditional NLP word representations (\eg term frequency-inverse document frequency), word embeddings are a far better alternative in the context of neural models, since it gives the optimization procedure the chance to learn better representations.
The seminal work of~\citet{bolukbasi2016man} of fairness in word embeddings describes how such vectorized word representations encapsulate societal biases.
Aimed with this knowledge, it became possible to use word-embedding manipulation techniques for bias mitigation in fairness-sensitive applications~\cite{bolukbasi2016man, bordia2019wordlevel}.
Most studies of word embedding debiasing follow a similar strategy to the Hard-Debiasing method~\cite{bolukbasi2016man}, where a ``protected'' subspace is identified and then words have their projection removed from that space.
Other word embedding debiasing approaches use a wide variety of helper tools, such as clustering~\cite{Du2020mdr}, a helper dictionary~\cite{kaneko2021dictionary}, causal inference~\cite{yang2020causal}, among others~\cite{aekula2021double, dev2021oscar, Vargas2020exploring, bolukbasi2016quantifying, kaneko2019gender, Gyamfi2020deb2viz, kumar2020nurse, dev2019attenuating, kumar2021hiperbolic, kaneko2021dictionary, lauscher2020general, gaci2022iterative}.

With the rise of Transformer-based architectures~\cite{vaswani2017attention}, which use (self-)attention as their primary inductive biases, methods that manipulate embedding spaces had to be adapted since attention uses original word-embedding representations to create \textit{contextualized} word representations.
That is, the semantic meaning of a token depends on the entire sequence of tokens.
A typical adaptation to work with contextual vector spaces is adding or complementing inputs with pre-defined sentence templates~\citep{kurita2019measuring}.
\citet{liang2020towards} use such templates to adapt the Hard-Debiasing approach to a contextual scenario.
Other studies rely on strategies like iterative generation to mitigate biases~\cite{liang2021towards,Subramanian2021evaluating}.
This type of method can be adapted to work on a wide range of pre-trained Transformer-based architectures, since they all rely on the attention mechanism.

Although vector-space manipulation is explored mainly in language applications, it is also possible to adapt the core ideas to CV.
\citet{salvador2022faircal} rely on clustering and embeddings to fix unfairness issues in the visual domain.
The authors propose a clustering-based calibration method to ensure that similar images have similar scores according to which cluster they belong to.
Also working with images, \citet{wang2021gender} remove the projection of CLIP embeddings from protected subspaces.

Much like other methods, vector-space manipulation approaches are not without drawbacks.
For instance, in the context of Transformer models, biases have different effects in each layer, adding considerable complexity to mitigation strategies~\cite{dev2020measuring, vig2020investigating}. 
Another difficulty is the necessity of a reference to guide the model to identify biased sub-spaces, which leads to the creation of repositories of biased and unbiased examples~\cite{bolukbasi2016man}.

\section{Challenges and Future Directions}
\label{sec:challenges-future-directions}

Throughout this survey, we presented several challenges regarding how to define the concepts of bias and fairness (Section~\ref{sec:bias-and-fairness}), how to evaluate ML models using individual or grouped notions of fairness (Section~\ref{sec:metrics}), and how to debias models to make them fairer (Section~\ref{sec:debias-methods}).
After analyzing the aforementioned issues, we now focus on general challenges for fairness in neural models while also highlighting potential future directions of research.
Specifically, we analyze
\begin{enumerate*}[label=(\roman*)]
    \item how to assess and present the fairness risks in neural models;
    \item how to identify who should be responsible for owning and dealing with fairness issues in neural models;
    \item how to deal with the fairness-accuracy trade-off; and
    \item current challenges and research directions for fairer Foundation Models.
\end{enumerate*}

\subsection{Risk Awareness}
Every machine learning model is a direct consequence of several design decisions, such as dataset collection and filtering, model architecture, optimization procedure, and hyperparameter selection.
Since these design decisions have a direct influence not only on the performance of a model but also on its perceived fairness, it is essential to clearly communicate these when internally or publicly releasing a model to highlight its limitations and intended usage.
That way, all relevant stakeholders can make informed decisions concerning the usage of specific technologies in the context where they operate.
However, deciding how and when to communicate such design decisions is not a trivial matter.
Thus, we now highlight recent advances and best practices to help ML researchers and practitioners create proper documentation to improve risk awareness.

\subsubsection{Artifact Documentation}
Several studies~\cite{bender2018data, benjamin2019towards, mitchell2019model, holland2020dataset, gebru2021datasheets} have attempted to standardize the communication of dataset and model design decisions in hopes of stimulating best practices in the creation of such artifacts, especially regarding sensitive topics such as fairness, privacy, and security.
\textit{Datasheets for datasets}~\cite{gebru2021datasheets} raises several questions (grouped into sections that roughly match the key stages of a dataset life cycle) to help dataset creators and consumers think about a dataset holistically, asking details on the motivation for creating the dataset, how the data was collected and aggregated, the recommended uses, and so on.
\textit{Datasheets} may help users to select more appropriate datasets for a specific task, increase transparency, mitigate unwanted societal biases, and even increase reproducibility.
There are several alternative approaches for documenting datasets; two examples include the Dataset Nutrition Label~\cite{holland2020dataset} and Data Statements for NLP~\cite{bender2018data}.
Critically, all of the aforementioned approaches highlight the importance of documenting datasets to inform end-users of fairness-related limitations since most biases are created or inherited in the data collection stage.

Regarding model documentation, \textit{Model Cards}~\cite{mitchell2019model} are an excellent protocol to increase transparency concerning training and evaluation protocols across different bias and fairness metrics (especially among different values of protected attributes, both individually and combined), and clarifying the intended use cases of models to minimize their usage in inappropriate contexts.
Model cards are especially important in research and open-source initiatives, since it leads to a more responsible and accountable model democratization process, which allows stakeholders to compare candidate models across not only traditional performance metrics (\eg accuracy for classifiers) but also ethical and fairness considerations.

Often, artifact creators must also legally protect themselves from misuse.
The Montreal Data License (MDL)~\cite{benjamin2019towards} is a project that attempts to improve the taxonomy of data licensing to better suit machine learning and artificial intelligence.
In the context of fairness, MDL provides optional restrictions regarding ethical considerations when granting rights to use and/or distribute a dataset, \eg restricting usage in high-risk scenarios such as health-related fields or military applications.

\subsubsection{Research Documentation}
Several mechanisms exist (\eg institutional review boards, conference program committees, and funding bodies) that provide the academic community with means to prevent unethical research.
However, there are few implemented protocols concerning fair research in mainstream deep learning conferences, and deep learning papers often do not discuss such limitations.
Since conference papers tend to be prioritized over journal publications in computer science (and deep learning especially), conference organizers should strive to improve fairness awareness in conference publications.

Some deep learning conferences have recently included protocols to help researchers communicate fairness constraints.
In 2020, NeurIPS required that \textit{all} papers include a ``broader impact statement'' covering the ethical and fairness aspects of research in their camera-ready versions~\cite{neurips2020callforpapers}
Additionally, the conference incorporated a new ethics review process where technical reviewers could flag papers for ethical concerns, which a pool of ethics reviewers would later analyze.
These initiatives had a mixed reception from the academic community due to their perceived political nature.
Several studies have analyzed the results of this experiment~\cite{prunkl2021institutionalizing, ashurst2022ai, nanayakkara2021unpacking} and arrived at the following conclusions:
\begin{enumerate*}[label=(\roman*)]
    \item although $10$\% of all papers opted out of writing the broader impact statement, this occurred mainly in theoretical subareas.
    Most authors took the opportunity to reflect on their work rather than stating that the statement was not applicable.
    \item the average statement length was $168$ words and the distribution presents a long tail ---the longest statement had $4000$ words.
    The areas of CV and NLP differ in average statement length ($166$ and $223$, respectively).
    \item authors focused more on positive societal impacts rather than negative ones, which intuitively goes against the main purpose of the experiment.
\end{enumerate*}

The lack of standardization and clear guidance regarding the expected contents of a broader impact statement and misaligned incentives for researchers may explain the results of this experiment.
However, since this was the first time that broader impact statements were mandatory, it is hard to judge the idea's true potential and how researchers would behave in the following years.
Following the feedback of the broader impact statement implementation, the NeurIPS conference opted to replace the statement with a ``Paper Checklist'' that provides researchers with a list of best practices for responsible machine learning research~\cite{neurips2021paperchecklist1, neurips2021paperchecklist2}.
This type of mechanism is an attempt to force researchers to think about a study's negative societal impacts before it is released, which may also increase risk awareness for ML practitioners who may use open-source code or attempt to reimplement algorithms.

\subsubsection{Fairness Frameworks}
One of the major goals of fairness research is the inclusion of fairness metrics and techniques into traditional ML pipelines to help mitigate biases.
Considerable efforts have been made to create open-source fairness software toolkits that help ML practitioners audit datasets and models~\cite{saleiro2018aequitas, bellamy2019ai, wexler2019if, tensorflow2020fairness, vasudevan2020lift, bird2020fairlearn, google2020fairness}.
These toolkits implement visualization and interpretability techniques, fairness metrics, and state-of-the-art debiasing techniques, translating research into actionable procedures.
However, there is a lack of application of such solutions in practice, especially in industry.
According to~\citet{richardson2021framework}, several factors explain this:
\begin{enumerate*}[label=(\roman*)]
    \item there are too many fairness metrics, and the differences between them are not clear to practitioners.
    Furthermore, major trade-offs exist between fairness metrics, and often it is mathematically impossible to optimize multiple fairness metrics.
    Choosing the ``right'' metric is delegated to practitioners, who are unfamiliar with the technical aspects of fairness research.
    
    \item there exists a disconnect between fairness research in academia and industry, which translates to a lack of applicability of procedures and metrics in industry settings.
    Additionally, fairness concepts are highly domain-dependant, differing substantially between domain applications, and yet most domains lack thorough and specific algorithmic bias guidance.

    \item frameworks often do not provide help with communicating fairness concerns and trade-offs to stakeholders, which inevitably reduces the adoption of frameworks due to organizational frictions.
\end{enumerate*}

Several suggestions have been proposed to mitigate these issues, and many involve improvements in communication between academics and practitioners.
\citet{friedler2016possibility} suggest that fairness experts clearly state the priorities of each fairness metric, while~\citet{verma2018fairness} suggest that researchers clarify which definitions are appropriate for which situations.
\citet{friedler2016possibility} argue that new fairness metrics should only be introduced if they behave fundamentally differently from existing metrics.
Several practitioners have also requested domain-specific procedures and metrics, and that fairness experts create knowledge bases for each domain~\cite{holstein2019improving}
Based on user feedback, fairness researchers should also consider creating taxonomies of potential harms and biases~\cite{madaio2020co, cramer2018assessing}, easy-to-digest summaries explaining biases and their potential sources~\cite{garciagarthright2018assessing, cramer2018assessing}, guidelines for best practices throughout the ML pipeline~\cite{holstein2019improving}, and tutorials exemplifying how to incorporate fairness~\cite{holstein2019improving, madaio2020co, richardson2021towards}.
We believe that organizations in industry settings also have a big part to play in ensuring the implementation of fairness protocols.
Organizations should promote a culture of fairness awareness and incorporate fairness as a global objective (like security, privacy, and accessibility)~\cite{garciagarthright2018assessing, madaio2020co, stark2019data}.
Additionally, organizations should strive to provide practitioners with resources and fairness support teams that provide knowledge and actionable steps in fairness issues during all steps of the ML pipeline~\cite{madaio2020co, mittelstadt2019principles}.

\subsubsection{Final Considerations}
We believe it is the responsibility of data and model creators to communicate fairness risks.
Artifact documentation may also be helpful for policy makers, investigative journalists, and individuals whose data are included in the datasets or who would be impacted by the deployment of such models.
We recommend that the language used in such documents range from accessible to technical, allowing both laypersons and domain experts to understand the critical decisions made to create a specific artifact.
However, none of the approaches presented in this section provide a complete solution for mitigating unwanted societal biases or potential risks accrued by the usage of \textit{any} algorithmic approach.
Society constantly changes, thus dataset and model creators will never anticipate every possible use of a particular artifact, and neither should they attempt to do so.
Instead, they should focus on objective facts and actions taken during the artifact creation pipeline, which will help consumers, companies, and governmental entities decide whether the technology is appropriate for their context and society as a whole.

\subsection{Risk Ownership}
An important but often overlooked discussion is deciding who is responsible for preventing algorithmic misuse.
Different scenarios require different levels of attention: systems that do not involve humans-in-the-loop (\eg autonomous driving or automated medical diagnosis) significantly increase the impact of algorithmic biases.
Sometimes, adjustments should be performed by the owners of the technology (not only developers but also organizations).
In other cases, however, the user of an AI-based system must be responsible for correctly using a tool or adjusting outputs to fit the notions of bias and fairness within the context in which they operate.
Ultimately, we believe that fairness risk ownership is tied to two concepts: how much control a user has over the model's outputs and whether the user is directly affected by model decisions.
To illustrate this challenge and provide some guidance, we contrast two exemplar situations of model deployment, namely facial recognition and image generation, where we believe that the responsibility of dealing with unfairness falls to artifact owners and users, respectively.

\subsubsection{Algorithm Owner}
A facial recognition system extracts facial features to verify a person's identity and has a wide range of applications.
Since facial recognition is one of the most popular biometric modalities due to its simple data collection process, several companies have invested in software-as-a-service facial recognition products and incorporated them into mainstream technological products, such as smartphones.
However, some studies~\cite{klare2012face, Buolamwini2018gender, raji2020saving} have analyzed the potential fairness pitfalls of neural facial recognition systems and concluded that they might discriminate based on protected attributes, such as race and gender, by performing significantly worse on specific demographics.
This discrepancy in performance is worrisome, especially considering that one of the clients of such facial recognition systems includes governments and law enforcement.
For instance, despite not directly determining the fate of an individual, such technology can be used to identify suspects in video surveillance footage, and erroneous misidentifications can have serious detrimental effects.
In this scenario, the ``victims'' of facial recognition systems have absolutely no control of the system's output and can be directly impacted by its decisions.
Thus, the ownership of fairness risks in this context must fall into the artifact's owners.

\subsubsection{Algorithm User}
Image generation is a research topic that can potentially disrupt several markets, such as art, design, fashion, retail, and many others.
One example of an image generator is DALL-E~\cite{ramesh2022hierarchical}, a text-to-image model created by OpenAI.
Since DALL-E's training dataset was collected from the Internet, it was expected that the model would inherit societal biases.
Regarding fairness, OpenAI implemented several measures to prevent algorithmic misuse~\cite{mishkin2022risks} but, despite these efforts, the company was still heavily criticized, and DALL-E was described as a ``harmful tool'' for reproducing societal biases, such as the (lack of) association of certain ethnicities with certain roles in society (\eg Black CEOs or female garbage collectors) while facilitating the creation of ``deep fakes''.
OpenAI effectively positioned itself as the owner of the risk of someone misusing their product, and all decisions of what constitutes an unfair generation became centralized at the company level.
However, OpenAI did not have to own this risk since users can control the model's output via input prompts and are not directly harmed by generated images.
Additionally, the model is indeed capable of generating diverse outputs, given the right prompts.
A more straightforward solution would be to provide disclaimers regarding fairness limitations and instructions on how to construct prompts that generate ``fairer'' and more diverse images and let users decide how they want to steer image generation since different solutions result in different trade-offs.

\subsection{Fairness-accuracy Trade-off}
As explored in Section~\ref{sec:metrics}, fairness researchers have already proposed several quantitative fairness metrics, giving rise to optimized objectives alongside task-specific metrics.
However, metrics usually either emphasize individual or group notions of fairness, and it is often mathematically impossible to optimize multiple fairness metrics~\cite{pessach2020algorithmic, chouldechova2017fair, barocas2019fairness}, which forces practitioners to select the most appropriate metric to optimize (which is context-dependant).
Additionally, it has been empirically observed that a trade-off exists between fairness and task performance and that increasing fairness often results in lower overall performance.
This gives rise to the challenge of analyzing the fairness-accuracy trade-off for a given scenario.
Practitioners must be careful when determining how to measure model performance, since this can be done in several ways and the choice of performance measure(s) can disguise or create new ethical concerns.
A reduction in accuracy may be the best outcome, especially if the difference in performance can be explained by algorithmic unfairness.

A potential solution to understand the trade-off characteristics between task and fairness metrics is to rely on advances in multi-task optimization literature to help neural models during optimization~\cite{crawshaw2020multi, vandenhende2021multi, gong2019comparison}.
One example of such a technique is generating a Pareto frontier to determine the set of Pareto-efficient solutions for a specific combination of fairness and task-specific metrics~\cite{martinez2020minimax, shah2021rawlsian, lin2019pareto, ruchte2021scalable, nia2022rethinking, haas2019price}.
Pareto efficiency corresponds to a situation where the performance of a model regarding a specific criterion cannot be made better without reducing the performance on at least one individual metric.
Another potential solution is to use techniques that increase fairness through proxies, such as disentanglement and causal inference, since they usually generalize better (although often at the cost of task performance).

\subsection{Fairness in Foundation Models}
A Foundation Model corresponds to ``any model that is trained on broad data at scale and can be adapted to a wide range of downstream tasks''~\cite{bommasani2021opportunities}.
FMs were initially introduced in the context of NLP research but are rapidly causing a revolution in all areas of deep learning research and industry applications.
The term ``foundation'' specifies the role of this category of models: a FM is incomplete by itself, but serves as the common basis from which many task-specific models are built via adaptation.
Since such models have the potential of being adapted to several tasks, biases may be perpetuated or amplified if not adequately addressed, making fairness a fundamental research direction for FMs~\cite{bommasani2021opportunities, meade2021empirical, delobelle2021measuring, schick2021self}.

Mitigating biases in FMs is not a trivial matter, especially due to their dataset and training regimes.
The datasets that support the training of FMs contain hundreds of millions or even billions of instances.
Thus applying pre-processing debiasing techniques is not an attractive option due to its potential impact on monetary cost and generalization capabilities.
One-step-training debiasing techniques should also be avoided since it is impossible to optimize for all fairness concepts simultaneously, and FMs are used in several scenarios and contexts.
For these reasons, the most promising research directions regarding bias mitigation in FMs are fine-tuning and prompting approaches~\cite{gira2022debiasing, sheng2020towards, berg2022prompt, gehman2020real, sharma2022sensitivetranslation, schick2021self} (Sections~\ref{sec:fine-tuning} and~\ref{sec:prompting}).
We also reiterate the importance of clearly communicating the risks of such technologies to prevent potential algorithmic misuse and highlight the importance of open-sourcing such models to accelerate the discovery of potential fairness issues.

\section{Conclusion}
In this survey paper, we have investigated debiasing methods targeting fairness-aware neural networks for language and vision research.
We have contextualized fairness and its relationship with biases and their possible origins.
We have presented the main metrics and evaluation measures for assessing the level of fairness provided by models for computer vision and natural language processing tasks, reviewing both application-specific and general-purpose measures, their proper use, applicability, and known limitations.
Then, we have discussed, in depth, several debiasing methods for neural models under the perspective of a new taxonomy for the area, which is yet another contribution of this paper.
We concluded with our thoughts on the most pressing fairness challenges in neural networks, calling attention for potential trends and future research directions.

We should point readers to Table~\ref{tab:debiasing-methods}, which facilitates the identification of research gaps and potential saturation for specific categories of methods.
Following our proposed taxonomy, certain  categories of methods have no proposed approaches for specific modality types, which could point to low-hanging fruits.
Considering the prevalence of Foundation Models and the fairness problems they currently present, we also urge readers to dedicate significantly more effort into debiasing large-scale models.

We hope this survey enables researchers to quickly understand the issues we are facing, and we stress that it is not enough simply providing new neural models without paying attention to the potential harms and consequences that such models may have on underprivileged groups or individuals.

\begin{acks}
This work was funded by Motorola Mobility Brazil.
\end{acks}

\bibliographystyle{ACM-Reference-Format}
\bibliography{_bibliography}

\newpage
\appendix

\end{document}